\documentclass[9pt, conference]{IEEEtran}
\IEEEoverridecommandlockouts
\usepackage{cite}
\usepackage{amsmath,amssymb,amsfonts}
\usepackage{graphicx}
\usepackage{textcomp}
\usepackage{xcolor}
\usepackage[ruled]{algorithm2e}
\usepackage{multirow}
\usepackage{subfigure}

\def\figref#1{Figure~\ref{#1}}
\def\secref#1{Section~\ref{#1}}

\def\tabref#1{Table~\ref{#1}}

\def\algoref#1{Algorithm~\ref{#1}}

\def\BibTeX{{\rm B\kern-.05em{\sc i\kern-.025em b}\kern-.08em
    T\kern-.1667em\lower.7ex\hbox{E}\kern-.125emX}}

\begin{document}

\title{HierarchyFL: Heterogeneous
Federated Learning  via Hierarchical 
Self-Distillation }

\author{
Jun Xia, Yi Zhang, Zhihao Yue,  Ming Hu, Xian Wei, and Mingsong Chen\\
\thanks{Jun Xia, Yi Zhang, Zhihao Yue,  Ming Hu, Xian Wei, and Mingsong Chen are with the MoE Engineering Research Center of Software/Hardware Co-Design Technology and Application, East China Normal University, Shanghai 200062, China. 
Mingsong Chen is the corresponding author (e-mail: mschen@sei.ecnu.edu.cn).
}
}

\maketitle

%  \author{
% Jun Xia$^1$\and
%  Ting Wang$^1$\and
%  Jiepin Ding$^1$\and
%  Xian Wei$^1$\And
%  Mingsong Chen$^1$
 
%  \affiliations
%  $^1$MoE Eng. Research Center of SW/HW Co-Design Tech. and App.,
%  East China Normal University\\
% % $^3$Third Affiliation\\
% % $^4$Fourth Affiliation
% % \emails
% % \{jxia, jpding\}@stu.ecnu.edu.cn, \{twang, xwei, mschen\}@sei.ecnu.edu.cn
% % % third@other.example.com,
% % % fourth@example.com
% }

\begin{abstract}

Federated learning (FL) has been recognized as a privacy-preserving distributed machine learning paradigm that enables knowledge sharing among various heterogeneous artificial intelligence (AIoT) devices through centralized global model aggregation. FL suffers from model inaccuracy and slow convergence due to the model heterogeneity of the AIoT devices involved. Although various existing methods try to solve the bottleneck of the model heterogeneity problem, most of them improve the accuracy of heterogeneous models in a coarse-grained manner, which makes it still a great challenge to deploy large-scale AIoT devices. To alleviate the negative impact of this problem and take full advantage of the diversity of each heterogeneous model, we propose an efficient framework named HierarchyFL, which uses a small amount of public data for efficient and scalable knowledge across a variety of differently structured models.  By using self-distillation and our proposed ensemble library, each hierarchical model can intelligently learn from each other on cloud servers. Experimental results on various well-known datasets show that HierarchyFL can not only maximize the knowledge sharing among various heterogeneous models in large-scale AIoT systems, but also greatly improve the model performance of each involved heterogeneous AIoT device.

\end{abstract}


\section{Introduction}
%%% Heterogeneous IoT systems
Along with the prosperity of Artificial Intelligence of Things (AIoT), more and more heterogeneous devices possess deep neural networks for the purpose on precising discrimination and perceptual control \cite{fedavg, chand_dac2022, gao_iccad2021, zhang_tcad2021}.
However, due to both the limitation of training data and the restricted hardware resources,  the daily performance of  models in heterogeneous devices cannot be guaranteed.    To enhance the model  performance of heterogeneous AIoT devices, we are expecting a large-scale AIoT system that fully uses the  computing power of cloud computing to predict inference results accurately.  Although such a cloud computing system is promising to deal with collaborative learning of heterogeneous AIoT devices in daily life, due to the uncertainty of the real environment, it inherently suffers from the risks of data privacy exposure and model performance inaccuracy.
%%%%%%%%%%%%%  FL and its shortcomings   %%%%%%%%%%%

As a promising technology in distributed machine learning, Federated Learning (FL) is usually deployed in many security-critical scenarios (e.g., autonomous driving, portable medical equipment, and intelligent transportation) to avoid the risk of sensitive  data leakage \cite{safety_dac2022, light_dac2021, kairouz2021advances, wu_iccad2021}.
%为了保护本地数据的隐私，联邦学习通过在AIOT系统中上传模型权重的梯度而不是上传本地数据，使用聚合后模型的梯度进行全局模型的更新从而保证数据隐私。
%由于联邦学习使用同构模型进行聚合，这种理想的聚合方式并不符合实际场景中多样物联网设备具备异构资源的特性。从而引起如联邦学习训练等待，模型收敛效果差等诸多问题,这将严重影响联邦学习在AIoT场景中的部署。因此，如何使用xxx变成了一个亟待解决的问题。
To achieve this goal, FL ensures data privacy by uploading the gradients of the DNN model  in an AIoT system instead of uploading its locally sensitive data.
However, the classic FL uses the same global model for communication, which ignores the heterogeneity of various heterogeneous AIoT devices \cite{xu_dac2021}.
This ideal assumption  suffers from poor model performance and slow convergence speed, which greatly limits the FL development
 in  large-scale AIoT systems. 
According to the above reason, \textit{how to mitigate its negative effects  of model heterogeneity and enhance the   knowledge
sharing among heterogeneous models is becoming a great challenge in the development of FL framework, especially when facing  a large-scale AIoT system. }

%%%%%%%%%%%%%  the model heterogeneity  work   %%%%%%%%%%%

To weaken the negative effects of  the model heterogeneity issue, 
various existing works (e.g., FedProto \cite{fedproto}, PervasiveFL \cite{xia_tcad2022}, HeteroFL \cite{iclr_diao2021}, and FedGen \cite{zhu_icml2021}) have been proposed. However, most of them focus on the perspectives of  uploading  model gradients or middle-ware knowledge, which ignores the diversity of heterogeneous models.
% However, the classic FL uses the same global model for communication, which does not take the heterogeneity of various AIoT devices into account.
To make this matter  worse, some of them focus overly on communication overhead, which has  a restricted effect on the enhancement performance of heterogeneous models.
For example, HeteroFL only considers sharing the same part of the global model and  neglects the diversity among the heterogeneous models of all AIoT devices, which greatly influences its quality of knowledge sharing between various heterogeneous models. 
%fed-prototype
To conceal the model heterogeneity for each AIoT device, PervasiveFL and Fedproto try to send a same ``modellet'' or ``prototype'' as its knowledge middle-ware to all the AIoT devices for effective knowledge sharing. 
To further extract the knowledge of heterogeneous models, FedGen enables the knowledge sharing by dint of a generator model, which introduced an extra communication overhead in FL. 
Although various existing methods try to tackle the bottleneck of the model heterogeneity problem, most of them improve the accuracy of heterogeneity models in a coarse-grained manner, which makes the deployment of  large-scale AIoT devices a great challenge. 

%%

%%%%%%%%%%%%%  our contribution %%%%%%%%%%%%%
To enable effective knowledge sharing between heterogeneous models in large-scale AIoT systems, this paper proposed a novel framework named \textit{HierarchyFL}, which allows different hierarchy models to learn from each other, thus improving their inference performance. 
This paper makes the following three major contributions: 
\begin{itemize}

% \item  We introduce the concept of 
% {\it modellet} that acts as an omnipotent portal for FL. 
\item  Inspired by  self-distillation method \cite{pami_self_distillation}, we propose the HierarchyFL, which allows the distillation between 
various hierarchy models on  the cloud server.  
Since  all the  hierarchy local models  (i.e., heterogeneous local models) in an AIoT system are  the subset of global models, they enable  knowledge sharing among
heterogeneous devices  by  layer-alignment averaging way with different hierarchies of the global model.

\item  
To fully utilize the diversity of each heterogeneous model in self-distillation, we propose the concept of \textit{meta-leaner} that can explore the proportion of each heterogeneous model in knowledge sharing. Based on the meta-learner, we develop an ensemble distillation library that  has  both the ensemble logits and ensemble features of all  hierarchy models to maximize the efficacy of self-distillation in knowledge sharing of various heterogeneous models.

\item We design serials of  experiments on two well-known datasets with both various Non-Independent and Identically Distributed  (Non-IID)  and IID settings. 
% Experimental results show that HierarchyFL can accommodate  large-scale AIoT systems  involving various hierarchy  models of an AIoT system. 
Compared with various state-of-the-art heterogeneous FL approaches, HierarchyFL cannot  achieve better model inference performance  but also have superior scalability in large-scale AIoT systems. 
\end{itemize}

The rest of this paper is organized as follows. \secref{rel} presents related work on heterogeneous federated learning. After \secref{app}
details our proposed HierachyFL method, \secref{exp}
introduces the experimental results on well-known benchmarks. Finally, \secref{con} concludes the paper.

\section{Related Work}
\label{rel}

% 1. 联邦学习 zxq， pervasiveFL
% 2.异构联邦（HeteroFL， Fedproto， pervasiveFL）
% 3. 原型优化，异构优化，普适优化
% To the best of our knowledge, our approach is the first xxx.

\begin{figure*}[ht] 
	\begin{center} 
		\includegraphics[width=0.78\textwidth]{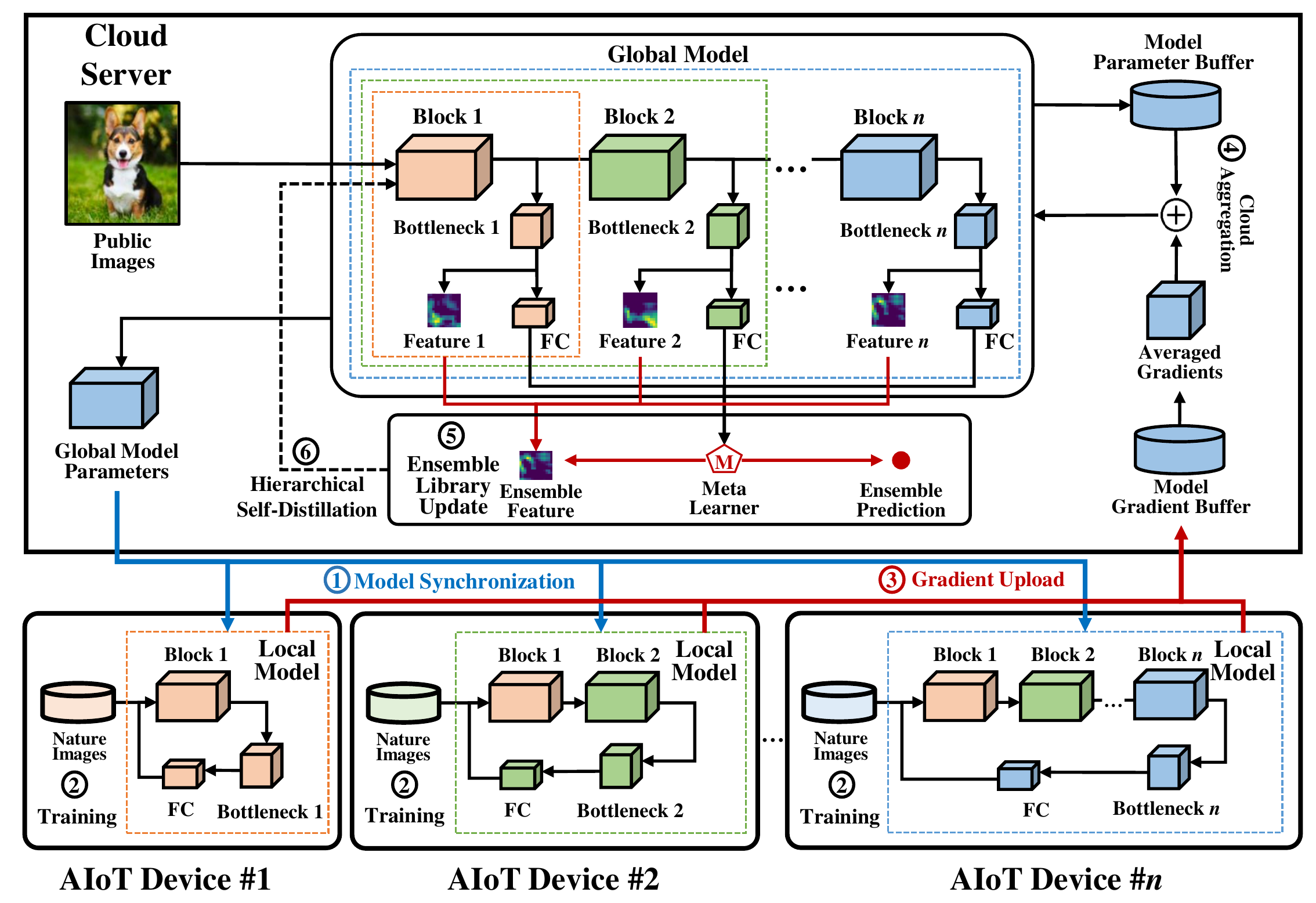}
		\caption{Framework and workflow of HierarchyFL}
		\label{fig:Architecture} 
	\end{center}
 \vspace{-0.2in}
\end{figure*} 

%尽管联邦学习目前已受到广泛研究，并被应用到各种领域，但它仍然遭受到客户端计算力不统一的问题。其中的一个解决方法是模型异质，通过根据客户端计算力部署相应大小的模型。大量的方法被提出用于提升模型异质的联邦学习性能，可以被分为基于蒸馏的方法[FedGen、PervasiveFL]、基于原型的方法[FedProto]、基于slim的方法（Slimmable-based methods）[HeteroFL、SplitMix]。 基于蒸馏的方法利用知识蒸馏实现异质模型信息的共享，如，FedGen在服务端通过集成蒸馏训练一个生成器共享全局知识，并在本地通过该生成器提升本地模型的性能。PervasiveFL通过在本地设置一个统一小的模型实现全局知识，利用深度互助学习解决联邦学习模型异质问题。基于原型的方法利用原型传递全局知识，打破了联邦学习模型传递知识的模式。FedProto让每个客户在训练过程中最小化本地原型与全局原型的差距。
%基于slim的方法的利用slim实现大模型到小模型的转变。HeteroFL对每个本地模型应用宽度瘦身，并在服务器端按通道聚合本地模型。Split-Mix提出利用训练瘦身模型并集成以获取大型模型。 

Although FL is promising in collaborative learning of AIoT systems, it still suffers from the  problem of model heterogeneity.
To mitigate the effect of this problem, various existing methods try to tackle  the bottleneck from the perspective of heterogeneous model knowledge sharing, which can  deploy a suitable heterogeneous model to an AIoT device. 
The methods for solving model heterogeneity can be divided into two categories, i.e., ``slimmable-based'' methods  and ``knowledge middleware'' distillation-based methods.

The former uses similar subset model aggregation to achieve knowledge sharing from large to small models.
For example,  Diao et al. proposed 
 a framework named HeteroFL \cite{iclr_diao2021}, which is applied width-wise  models and aggregates local models by channel on the cloud server. 
To fully use the heterogeneous models,  Hong et al. presented a novel method named split-Mix \cite{split_mix} that uses a thinner model in the cloud aggregation to obtain a more large model. 
Although the above methods are promising in dealing with the heterogeneous model problem, most of them still suffer from low inference accuracy of the heterogeneous model, which hinders their deployment in real scenarios.

To enable the effective knowledge sharing of all involved heterogeneous models, the latter uses the knowledge distillation method with the same knowledge middleware among heterogeneous models \cite{feddf}. For example,
to achieve effective knowledge sharing, Zhu et al. \cite{zhu_icml2021} proposed a distillation-based method named FedGen, which  designs a generator on the cloud server. Based on this generator, FedGen can not only share global knowledge through knowledge distillation but also enhance the inference performance
of local models.
To enable privacy-preserving FL for large-scale AIoT devices equipped with heterogeneous DNN models, Xia et al. \cite{xia_tcad2022} developed a novel framework named PervasiveFL, which uses deep mutual learning by setting up a uniformly small model named ``modellet'' to achieve global knowledge-sharing. 
Similar to PervasiveFL, Tan et al. \cite{fedproto} proposed a newly method named FedProto, which allows each device to send a prototype that represents the   features of each category in local nature images. In this way,  Fedproto can be used to minimize the gap between local and global prototypes to enhance knowledge sharing during training. 
Although the above methods are effective in improving
FL performance, most of them are based on coarse-granularity features.  None of them explicitly take the diversity of local heterogeneous
models to the model performance enhancement into account.

To the best of our knowledge, HierarchyFL is the first attempt to utilize  the merit of self-distillation and fully consider the diversity of local heterogeneous
models to mitigate the negative effect of model heterogeneity issues on AIoT devices.  Compared with state-of-the-art heterogeneous FL methods, HierarchyFL cannot only enable effective knowledge sharing between heterogeneous models in  large-scale AIoT systems but also  have superior
scalability in large-scale AIoT systems.

% and the deployment of Hierarchy is suitable in large-scale systems.

\section{Our HierarchyFL Approach}
\label{app}

To ease  the understanding of our HierarchyFL framework, 
this section introduces  our HierarchyFL architecture in detail. Firstly,  it  presents the architecture and workflow of HierarchyFL. 
Afterward,  it represents    hierarchical self-distillation procedure
based on our proposed ensemble library used in HierarchyFL. 
Lastly, it demonstrates
the training procedure of HierarchyFL on both the cloud server and  local devices.

\subsection{Architecture and Workflow  of HierarchyFL}

\subsubsection{Architecture   of HierarchyFL}
Typically, an AIoT application comprises a cloud server and various heterogeneous AIoT devices, where each  device has both  restricted memory and computation capacities.
In this paper, HierarchyFL focuses on the hierarchy model performance enhancement, which can not only enable knowledge sharing among heterogeneous models in large-scale AIoT systems but also greatly enhance the model performance in each AIoT device. 
As shown in \figref{fig:Architecture}, 
the HierarchyFL is composed of two components, i) the cloud server that  contains  the hierarchical self-distillation among various hierarchy models and the aggregation of each hierarchy model; ii) heterogeneous AIoT devices that conduct the local training based on the local nature images.
Unlike Fedavg, which uses the same global model in its training procedure, 
HierarchyFL allows  heterogeneous  models   to  adapt to the hardware resource capacities of AIoT devices.
To complete this idea, HierarchyFL  uses the layer heterogeneous (i.e., the layer width of all AIoT device models is variable.) models in an AIoT system.
To determine the size of the hierarchy model in each AIoT device, the device needs to upload its size of free hardware resources to the cloud server. 
Afterward, the cloud server can ensure and dispatch the  local  model in the newly participated devices for knowledge sharing.
\subsubsection{Workflow   of HierarchyFL}
In HierarchyFL, distributed
edge devices and a cloud server work closely to realize the global
learning of various kinds of hierarchy models deployed on heterogeneous
devices. Before training, all the devices participating in HierarchyFL will install and initialize a hierarchy model in advance, which is a subset of the global model in the cloud server. Afterward, 
the cloud server sends a part of the global model to  AIoT devices for local training.
When the local training end, HierarchyFL will conduct the model aggregation and hierarchical self-distillation on the cloud server.   Note that HierarchyFL allows hot-plug AIoT devices, where the newly involved device only inherits the parameters of the global model in the cloud server. In practical terms, the whole
working procedure of the HierarchyFL framework can be divided
into six steps.

{\bf Step 1 (Model Synchronization):} 
At the initial stage of HierarchyFL, the cloud server dispatches the part of global model parameters to each heterogeneous AIoT device 
according to their memory and computation capability for local training.

{\bf Step 2 (Local Training):} 
Based on the received global model parameters, each heterogeneous AIoT device constructs an initial hierarchy model (i.e., local model), which is trained by local nature images using cross-entropy loss to get the gradients of the local model for gradient upload.

{\bf Step 3 (Gradient Upload):} 
When the local training is finished,  each participating AIoT device uploads its latest local model gradients to a gradient buffer in the cloud server. All the collected local model gradients derive an aggregated global model by using layer-alignment averaging, where all the heterogeneous models only aggregate their corresponding public parts in a global model in an average way.

{\bf Step 4 (Cloud Aggregation):} 
After receiving the local model gradients of all the participating devices, this step will layer-alignment averaging such gradients and use the previous round global model stored in the model parameter buffer to construct a new global model. 

{\bf Step 5 (Ensemble Library Update):} 
After the global model  is averaged, this step will update the meta-learner by calculating the cross-entropy loss between the ensemble predictions of each hierarchy model and the labels of public images, thus utilizing the diversity of each hierarchy model. Based on the new meta-learner,  the   features and  predictions in the ensemble library can be updated for the  hierarchical self-distillation period. 
% Based on the public data in the cloud server, the meta-learner can be trained to get the importance of each hierarchy model in knowledge sharing. 
(See details in \secref{ensemble library update}). 

{\bf Step 6 (Hierarchical Self-distillation):} 
Based on the new ensemble feature and prediction, all  of the hierarchy models can be updated by hierarchical self-distillation,  which includes two categories, i.e., feature distillation and prediction distillation. Afterward, the updated global model can be saved to the global model parameters.

HierarchyFL repeats all the above six steps until the convergence of both global models and all  its hierarchy models. 
\subsection{Hierarchical Self-distillation}
In HierarchyFL,  the cloud server and all the participating AIoT devices 
start knowledge sharing by hierarchical self-distillation that fully considers the diversity of each hierarchy model. 
% To fully utilize the diversity of each heterogeneous model in self-distillation,
The hierarchical self-distillation procedure can be  aided by an ensemble library, which consists of three parts, i.e.,  meta-learner, ensemble predictions, and ensemble features. 
To explore the proportion of each heterogeneous model in knowledge sharing,
 the meta-learner can be updated by cross-entropy loss function between the ensemble predictions of each hierarchy model and the labels of public images based on a  few public data in the cloud server. Afterward, the ensemble library that includes ensemble features and ensemble predictions can be updated through the newly meta-learner.
When the ensemble library is updated, it can be used for  hierarchy self-distillation for each hierarchy model on the cloud server.

%由于受到自蒸馏方法的启发，本文的方法通过建立蒸馏库将每个层次模型的输出建立老师特征与预测集。基于公共全集数据，为了将本文中的每一个特征进行提取
\subsubsection{Ensemble Library Update}
\label{ensemble library update}
Inspired by self-distillation methods, the approach in this paper builds the ensemble features and prediction sets from the output of each hierarchical model by maintaining an ensemble  library for hierarchical self-distillation.
Assume that there are $n$ categories of hierarchy models  participating in an AIoT system, and let the meta-learner and the prediction of each hierarchy model be $M$ and $P$. Based on the few public images, the loss function of the meta-learner is demonstrated as follows
\begin{equation}
\footnotesize
L_{M} = CE\_Loss( \sum_{i=1}^{n} {M_{i} \cdot P_{i}} ,y).
\label{equ:1}
\end{equation}
where  $y$ is the labels of public data on the cloud server, $CE\_Loss$ means the cross entropy loss function of the meta-learner, and $M_{i}$ represents the proportion of $i^{th}$ hierarchy model.
Based on the newly meta-learner, the ensemble features $F_{M}$ and the ensemble predictions $P_{M}$ can be updated as follows
\begin{equation}
\footnotesize
P_{M} = \sum_{i=1}^{n} {M_{i} \cdot P_{i}}, \qquad 
F_{M} = \sum_{i=1}^{n} {M_{i} \cdot F_{i}}.
\label{equ:2}
\end{equation}
Here, $F_{i}$ is the features of $i^{th}$ hierarchy model, where it is a tensor before its fully connected layer.

\begin{algorithm}[htbp] 
\LinesNumbered %要求显示行号

	\KwIn{
\romannumeral1) $N$, number of devices;\\
\romannumeral2) $R$, number of communication rounds;\\
\romannumeral3) $C$, number of self-distillation epochs\;\\
\romannumeral4) $K$, number of hierarchy model types\;\\
\romannumeral5) $M$, the meta-learner in the cloud server\;\\

 	}
\While{ $true$}{
    \For{ $r  \leftarrow 1$ to $R$}{
        \For{$n  \leftarrow 1$ to $N$}
            {
            $\nabla w_{G}^{r, n}\ \ \leftarrow\ Receive(n)$\;  
            }
         $\nabla w_{G\ }^{r}\leftarrow layer\_ wise\_averaging(\nabla w_{H}^{r, n})$;\
         
         $w_{G} \leftarrow SGD.get\_weights (w_{G},\ \nabla w_{G\ }^{r})$\;

          \For{$c\leftarrow 1$ to $C$}
          {
            $(x_{public},\ y_{public})\leftarrow Public\_Data()$\;
            $\bigcup_{i=1}^{K}\left \{F_{i}, P_{i}\right \}$ $\leftarrow\ w_{G}\left ( x_{public} \right ) $\;
            $L_{M} \leftarrow CE\_loss$( $\sum_{i=1}^{n} {M_{i} \cdot P_{i}}$, $y_{public}$)\;
            $\nabla M \leftarrow SGD.get\_gradients (L_{M},\ w_{G})$\;
            $M \leftarrow Update(\nabla M, M$)\;
            $P_{M} \leftarrow \sum_{i=1}^{n} {M_{i} \cdot P_{i}}$ \;
            $F_{M} \leftarrow \sum_{i=1}^{n} {M_{i} \cdot F_{i}}$ \;
            $L_{G} \leftarrow loss\_calculation()$; (See Equation~\ref{equ:3})
            
            $\nabla w_{G} \leftarrow SGD.get\_gradients (L_{G},\ w_{G})$\;
            $w_{G} \leftarrow Update(\nabla w_{G}, w_{G}$)\;
            
         }
         
         \For{$n\leftarrow 1$ to $N$}
          {
            $Send(n, w_{G}^{n})$\;
         }
        
    }
    }			
 \caption{ Training Procedure of HierarchyFL (Cloud)}
	\label{alg:1}
\end{algorithm}

\begin{table*}[h]
\centering
% \scriptsize

  \caption{Test accuracy comparison for four scenarios using three heterogeneous federated learning methods.}
  \label{test_acc}
\begin{tabular}{|c|cccccccccccc|}
\hline
DataSet      & \multicolumn{12}{c|}{CIFAR10}                                                                                                                                                                                                                                                                                                                                                  \\ \hline
Methods      & \multicolumn{4}{c|}{HeteroFL}                                                                                              & \multicolumn{4}{c|}{Fedproto}                                                                                   & \multicolumn{4}{c|}{HierarchyFL (Ours)}                                                                                                 \\ \hline
Distribution & \multicolumn{1}{c|}{$\alpha$=0.1} & \multicolumn{1}{c|}{$\alpha$=0.5}          & \multicolumn{1}{c|}{$\alpha$=1.0} & \multicolumn{1}{c|}{IID}   & \multicolumn{1}{c|}{$\alpha$=0.1} & \multicolumn{1}{c|}{$\alpha$=0.5} & \multicolumn{1}{c|}{$\alpha$=1.0} & \multicolumn{1}{c|}{IID} & \multicolumn{1}{c|}{$\alpha$=0.1}          & \multicolumn{1}{c|}{$\alpha$=0.5}          & \multicolumn{1}{c|}{$\alpha$=1.0}          & IID            \\ \hline
Model 1      & \multicolumn{1}{c|}{49.40} & \multicolumn{1}{c|}{75.03}          & \multicolumn{1}{c|}{81.28} & \multicolumn{1}{c|}{85.86} & \multicolumn{1}{c|}{21.81} & \multicolumn{1}{c|}{49.89} & \multicolumn{1}{c|}{54.83} & \multicolumn{1}{c|}{72.03}    & \multicolumn{1}{c|}{\textbf{69.43}} & \multicolumn{1}{c|}{\textbf{81.33}} & \multicolumn{1}{c|}{\textbf{83.52}} & \textbf{86.41} \\ \hline
Model 2      & \multicolumn{1}{c|}{63.41} & \multicolumn{1}{c|}{79.46}          & \multicolumn{1}{c|}{84.10} & \multicolumn{1}{c|}{87.12} & \multicolumn{1}{c|}{25.05} & \multicolumn{1}{c|}{42.76} & \multicolumn{1}{c|}{61.80} & \multicolumn{1}{c|}{72.51}    & \multicolumn{1}{c|}{\textbf{76.91}} & \multicolumn{1}{c|}{\textbf{84.25}} & \multicolumn{1}{c|}{\textbf{86.00}} & \textbf{87.80} \\ \hline
Model 3      & \multicolumn{1}{c|}{67.93} & \multicolumn{1}{c|}{\textbf{87.80}} & \multicolumn{1}{c|}{88.17} & \multicolumn{1}{c|}{88.86} & \multicolumn{1}{c|}{30.24} & \multicolumn{1}{c|}{53.50} & \multicolumn{1}{c|}{64.52} & \multicolumn{1}{c|}{73.54}    & \multicolumn{1}{c|}{\textbf{83.18}} & \multicolumn{1}{c|}{86.86}          & \multicolumn{1}{c|}{\textbf{89.07}} & \textbf{89.70} \\ \hline
Model 4      & \multicolumn{1}{c|}{75.33} & \multicolumn{1}{c|}{\textbf{87.75}} & \multicolumn{1}{c|}{88.90} & \multicolumn{1}{c|}{89.21} & \multicolumn{1}{c|}{39.20} & \multicolumn{1}{c|}{58.59} & \multicolumn{1}{c|}{67.38} & \multicolumn{1}{c|}{73.22}    & \multicolumn{1}{c|}{\textbf{82.90}} & \multicolumn{1}{c|}{86.92}          & \multicolumn{1}{c|}{\textbf{89.15}} & \textbf{89.62} \\ \hline\hline
DataSet      & \multicolumn{12}{c|}{CIFAR100}                                                                                                                                                                                                                                                                                                                                                  \\ \hline
Methods      & \multicolumn{4}{c|}{HeteroFL}                                                                                              & \multicolumn{4}{c|}{Fedproto}                                                                                   & \multicolumn{4}{c|}{HierarchyFL (Ours)}                                                                                                 \\ \hline
Distribution & \multicolumn{1}{c|}{$\alpha$=0.1} & \multicolumn{1}{c|}{$\alpha$=0.5}          & \multicolumn{1}{c|}{$\alpha$=1.0} & \multicolumn{1}{c|}{IID}   & \multicolumn{1}{c|}{$\alpha$=0.1} & \multicolumn{1}{c|}{$\alpha$=0.5} & \multicolumn{1}{c|}{$\alpha$=1.0} & \multicolumn{1}{c|}{IID} & \multicolumn{1}{c|}{$\alpha$=0.1}          & \multicolumn{1}{c|}{$\alpha$=0.5}          & \multicolumn{1}{c|}{$\alpha$=1.0}          & IID            \\ \hline
Model 1      & \multicolumn{1}{c|}{34.85} & \multicolumn{1}{c|}{48.65}          & \multicolumn{1}{c|}{51.19} & \multicolumn{1}{c|}{54.88} & \multicolumn{1}{c|}{13.28} & \multicolumn{1}{c|}{21.23} & \multicolumn{1}{c|}{24.41} & \multicolumn{1}{c|}{31.08}    & \multicolumn{1}{c|}{\textbf{42.33}} & \multicolumn{1}{c|}{\textbf{52.32}} & \multicolumn{1}{c|}{\textbf{53.14}} & \textbf{56.31} \\ \hline
Model 2      & \multicolumn{1}{c|}{39.15} & \multicolumn{1}{c|}{50.96}          & \multicolumn{1}{c|}{53.84} & \multicolumn{1}{c|}{56.58} & \multicolumn{1}{c|}{14.84} & \multicolumn{1}{c|}{23.52} & \multicolumn{1}{c|}{25.28} & \multicolumn{1}{c|}{30.58}    & \multicolumn{1}{c|}{\textbf{46.44}} & \multicolumn{1}{c|}{\textbf{53.89}} & \multicolumn{1}{c|}{\textbf{56.26}} & \textbf{58.63} \\ \hline
Model 3      & \multicolumn{1}{c|}{45.13} & \multicolumn{1}{c|}{56.11}          & \multicolumn{1}{c|}{56.50} & \multicolumn{1}{c|}{57.55} & \multicolumn{1}{c|}{15.80} & \multicolumn{1}{c|}{24.05} & \multicolumn{1}{c|}{26.59} & \multicolumn{1}{c|}{30.06}    & \multicolumn{1}{c|}{\textbf{51.77}} & \multicolumn{1}{c|}{\textbf{58.95}} & \multicolumn{1}{c|}{\textbf{58.96}} & \textbf{60.16} \\ \hline
Model 4      & \multicolumn{1}{c|}{47.09} & \multicolumn{1}{c|}{57.64}          & \multicolumn{1}{c|}{57.29} & \multicolumn{1}{c|}{58.56} & \multicolumn{1}{c|}{18.18} & \multicolumn{1}{c|}{26.05} & \multicolumn{1}{c|}{27.00} & \multicolumn{1}{c|}{29.60}    & \multicolumn{1}{c|}{\textbf{52.56}} & \multicolumn{1}{c|}{\textbf{59.92}} & \multicolumn{1}{c|}{\textbf{59.49}} & \textbf{60.35} \\ \hline
\end{tabular}
\end{table*}

\subsubsection{Hierarchical Distillation}
After the ensemble library is updated, the ensemble features and predictions can be used as the ensemble knowledge for the hierarchical self-distillation, which can be fully considered  the diversity of each hierarchy model.
Note that since the ensemble library can get excellent performance  based on the diversity of each hierarchy model,  the distillation target of each hierarchy model should be different for effective knowledge sharing. 
To maintain the diversity of each hierarchy model, the loss function of hierarchical self-distillation can be formulated as 

\begin{equation}
\footnotesize
L_{G} = \sum_{i=1}^{n}
  \begin{cases}
      KL( P_{i}, P_{n}) + \beta \cdot MSE(F_{i}, F_{n}),   \quad \ if \quad  i <  n,  \\ 
	 KL( P_{i}, P_{M}) + \beta \cdot MSE(F_{i}, F_{M}),  \ if \quad i = n.
  \end{cases}
\label{equ:3}
\end{equation}
where $G$ is the global model in the cloud hierarchical self-distillation period, $KL$ and $MSE$ are the Kullback-Leibler  divergence and Mean Squared Error loss functions,  $\beta$  means the importance of $MSE$  loss function in hierarchical self-distillation. Based on this loss function,  each hierarchy  model can be trained from the ensemble library to realize effective knowledge sharing by hierarchical self-distillation.

\subsection{ Training Procedure of HierarchyFL}
%我们的方法分为两部分，一部分为云端微调操作，一部分为本地训练操作。其中，我们的云端服务器操作包含两步，一部分为按层聚合操作，由于模型的大小不同，层次聚合操作仅仅聚合与全局模型中相同的公有部分。在完成聚合后，为了将各个层次模型的优点进行利用，我们首先将meta-predictor进行训练，这
% This subsection introduces the implementation of HierarchyFL on
% devices and cloud, respectively. Algorithm 1 describes the  training procedure of HierarchyFL on the cloud server, and Algorithm 2 demonstrates the
% training  procedure of HierarchyFL on each AIoT device.
This subsection details the training procedure  of HierarchyFL on both devices and the cloud server.  
%For more details, please refer to algorithms \ref{alg:1} and \ref{alg:local training}. 

%由于每个

%基于云端的部分公有数据集，通过训练集成模型，我们获得了每个子层次模型的权重

\subsubsection{Training Procedure of HierarchyFL (Cloud)}

\algoref{alg:1} shows the training procedure of the cloud server in HierarchyFL, including gradient average, layer-wise aggregation, and hierarchical self-distillation. In \algoref{alg:1}, the training procedure 
 will be conducted with consecutive $R$ communication rounds in the cloud server (lines 2-23).   
Specifically, lines 3-5 receive gradients from $N$ heterogeneous AIoT deceives.
Line 6 conducts the   layer-alignment averaging way for gradients of each hierarchy model.
Line 7 aggregates the global model by adding the averaged hierarchy model gradients to the global model updated by the previous communication round. 
Lines 8-15 update the meta-learner based on a few public images, where  the ensemble features and predictions can be  updated by newly meta-learner.  Lines 16-19 use the updated ensemble library to execute hierarchical self-distillation for each hierarchy model.
Lines 20-22 dispatch the part of the global  model (i.e., the hierarchy model of all AIoT devices)  to all the participating AIoT devices in an asynchronous way.

\subsubsection{Training Procedure of HierarchyFL (Device)}

\begin{algorithm}[htbp] 
\label{algorithm:device}
\LinesNumbered %要求显示行号
% \footnotesize
\KwIn{
    \romannumeral1) $S$, cloud server;
    
    \romannumeral2) $n$, index of device;
    
    \romannumeral3) $R$, number of communication rounds;
    
    \romannumeral4) $E$, number of local training epochs;
    
    \romannumeral5) $D$, local device model;
    }
\KwOut{
$w_{D}$
}
    $(x,\ y)\leftarrow Collect()$\;
      \For{$r  \leftarrow$ $1$ to $R$}{  
    \For{$e  \leftarrow$ $1$ to $E$}{
            $ \left ( V^{D}, {P}^{D}\right ) \leftarrow\ w_{D}(x)$\; 
$L^{D} \leftarrow CE\_Loss (y,\ V^{D})$\;

            $\nabla w_{D}^{e} \leftarrow SGD.get\_gradients (L^{D},\ w_{D})$\;
            $w_{D} \leftarrow Update(\nabla w_{D}^{e}, w_{D}$)\;

    }
            $Send (\nabla w_{D}, n, S)$\;
            $w_{D} \leftarrow Receive ()$\;
    }

	\caption{ Training Procedure of HierarchyFL (Device)}
	\label{alg:local training}
\end{algorithm}

Algorithm~\ref{alg:local training} presents the training procedure of  our HierarchyFL in AIoT devices.
In line 1, 
each AIoT device gathers nature images  to support local training, where $x$ and  $y$ represent the gathered training data and their corresponding labels. 
Lines 2-11 conduct $R$ communication rounds of FL with the cloud server for the
 model gradients upload. Lines 3-4  utilize the nature images to   local model $w_{D}$  and figure out the corresponding labels of nature images from model predictions (i.e.,  $V^{D}$).
Lines 5-7 iteratively update the parameters of heterogeneous local models based on the cross-entropy loss function. 
After that, each local model uploads model gradients to the cloud server for aggregation.
 Based on  the trained  model, line 9 uploads the model gradient to the cloud server for aggregation.
 Finally, line 10 receives the newly dispatched model for the next local training round.

\section{Experiment}
\label{exp}
% RQ1: Performance of IID and non-IID scenarios

% RQ2: Scalibility analysis

% RQ3: Ablation Study

\begin{figure*}[ht]
\begin{center}
    \subfigure[\scriptsize{  CIFAR10 ($\alpha=0.1$) w/ 40 devices}]{\includegraphics[width=0.24\linewidth]{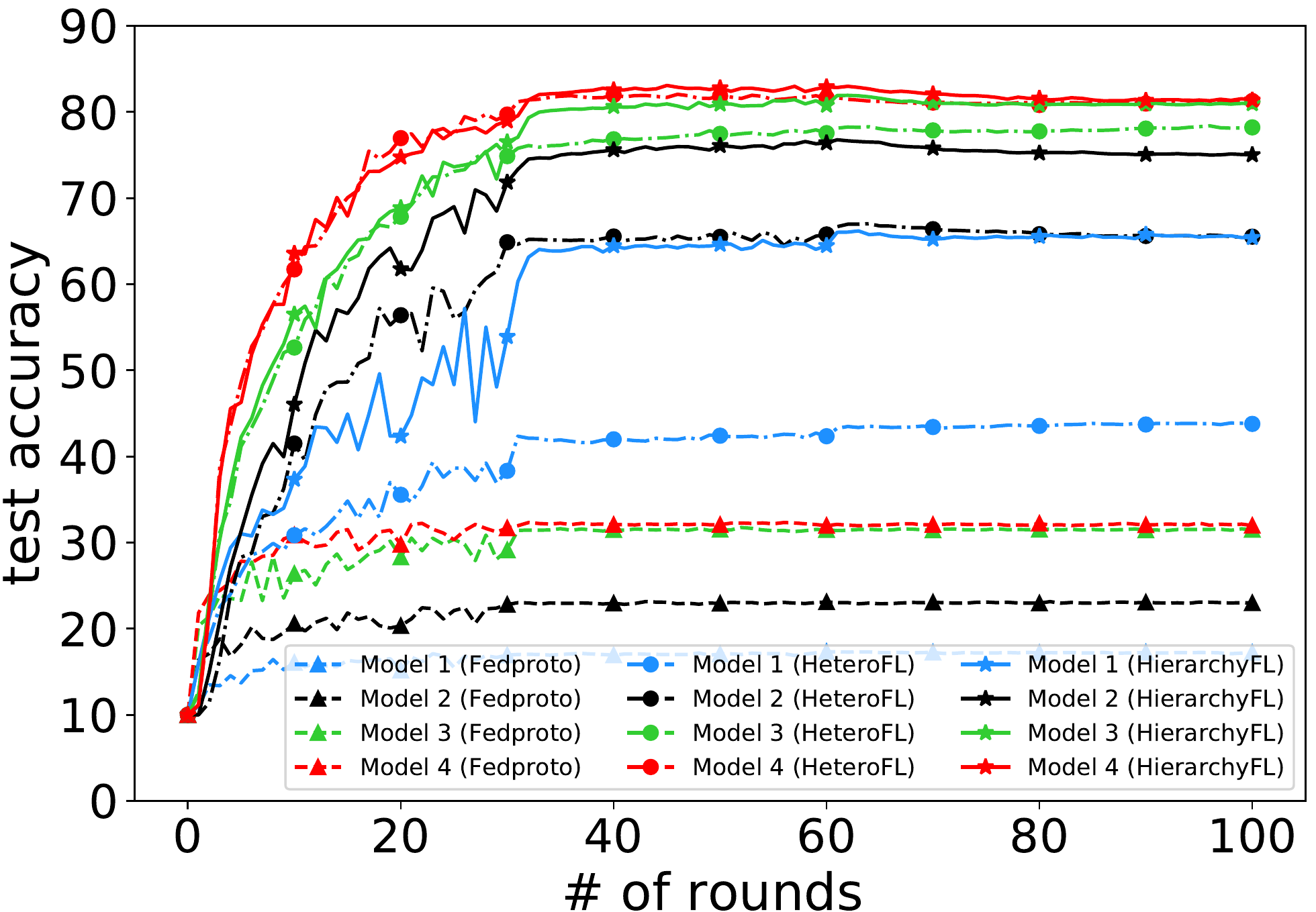}}
\subfigure[\scriptsize{ CIFAR10 ($\alpha=0.5$) w/ 40 devices}]{\includegraphics[width=0.24\linewidth]{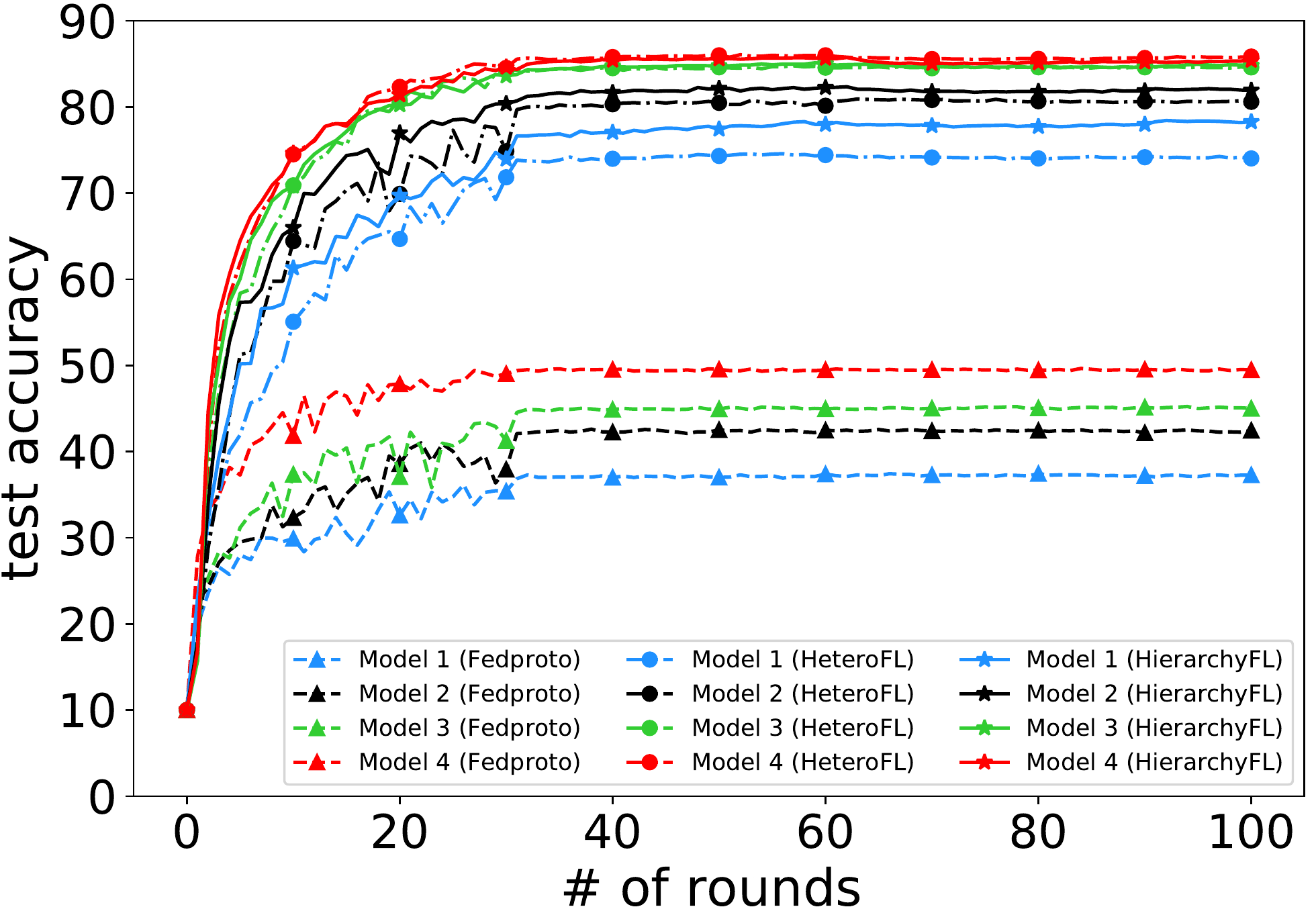}}
    \subfigure[\scriptsize{ CIFAR10 ($\alpha=1$) w/ 40 devices}]{\includegraphics[width=0.24\linewidth]{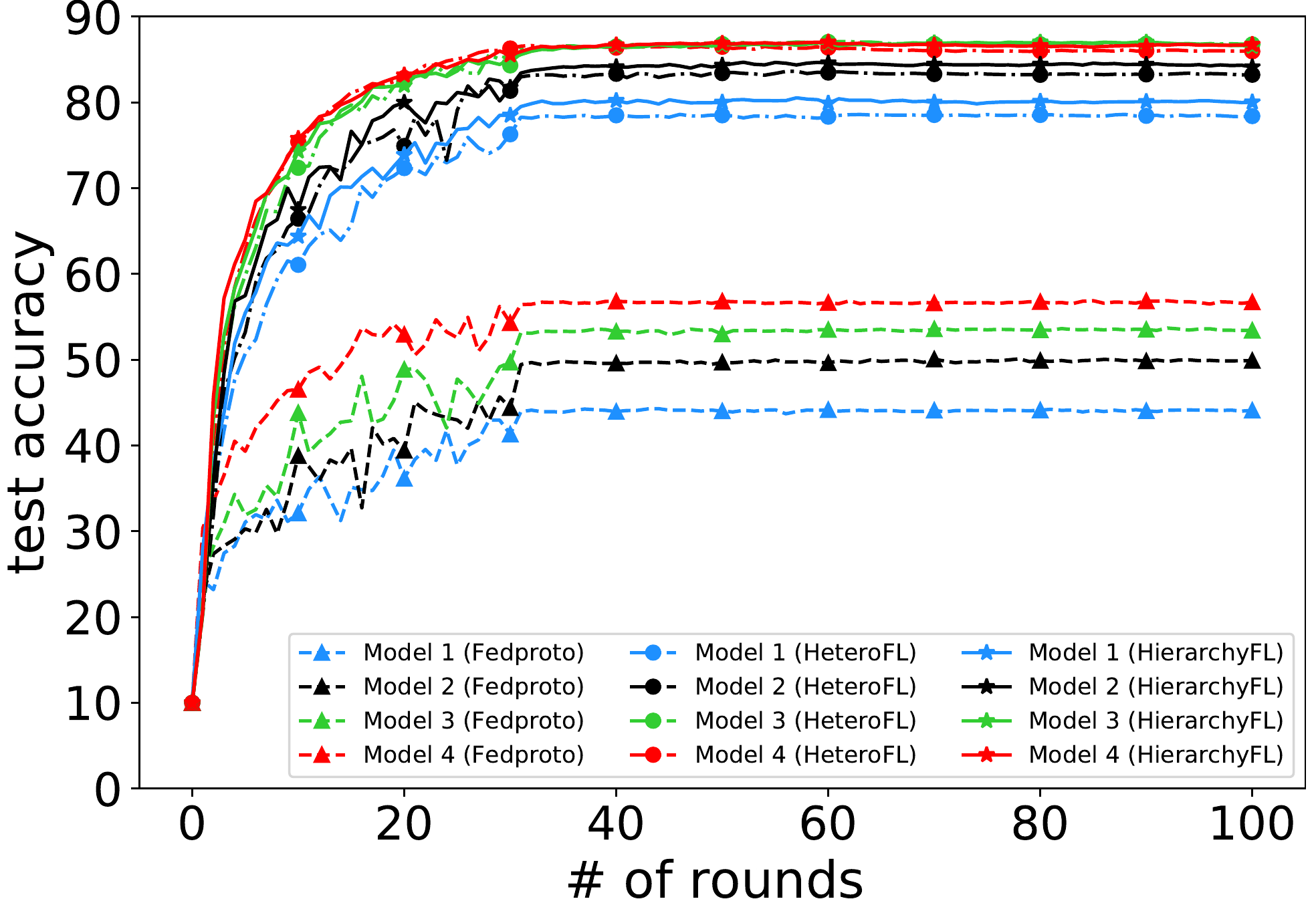}}
\subfigure[\scriptsize{ CIFAR10 (IID) w/ 40 devices}]{\includegraphics[width=0.24\linewidth]{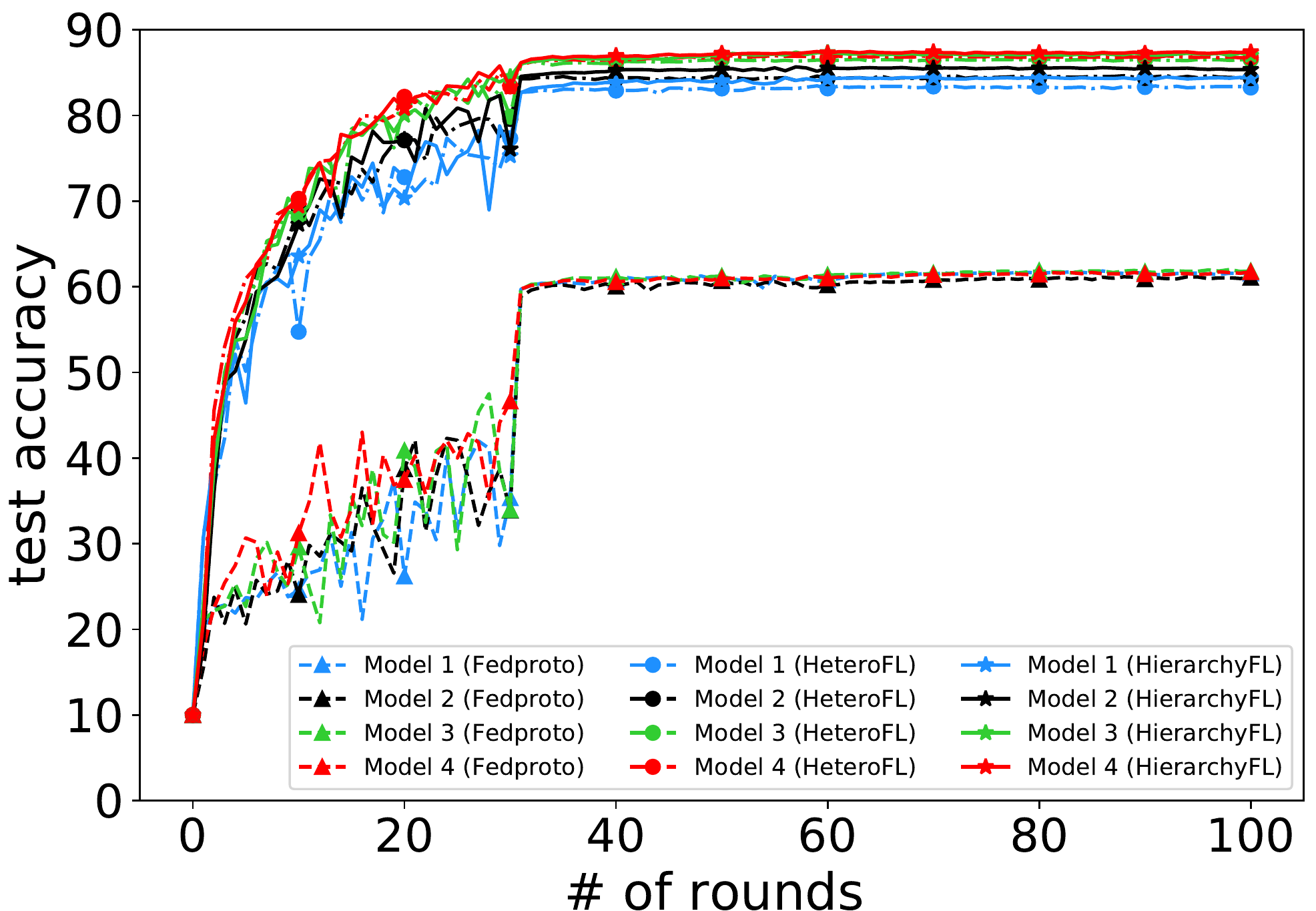}}
    \subfigure[\scriptsize{ CIFAR100 ($\alpha=0.1$) w/ 40 devices}]{\includegraphics[width=0.24\linewidth]{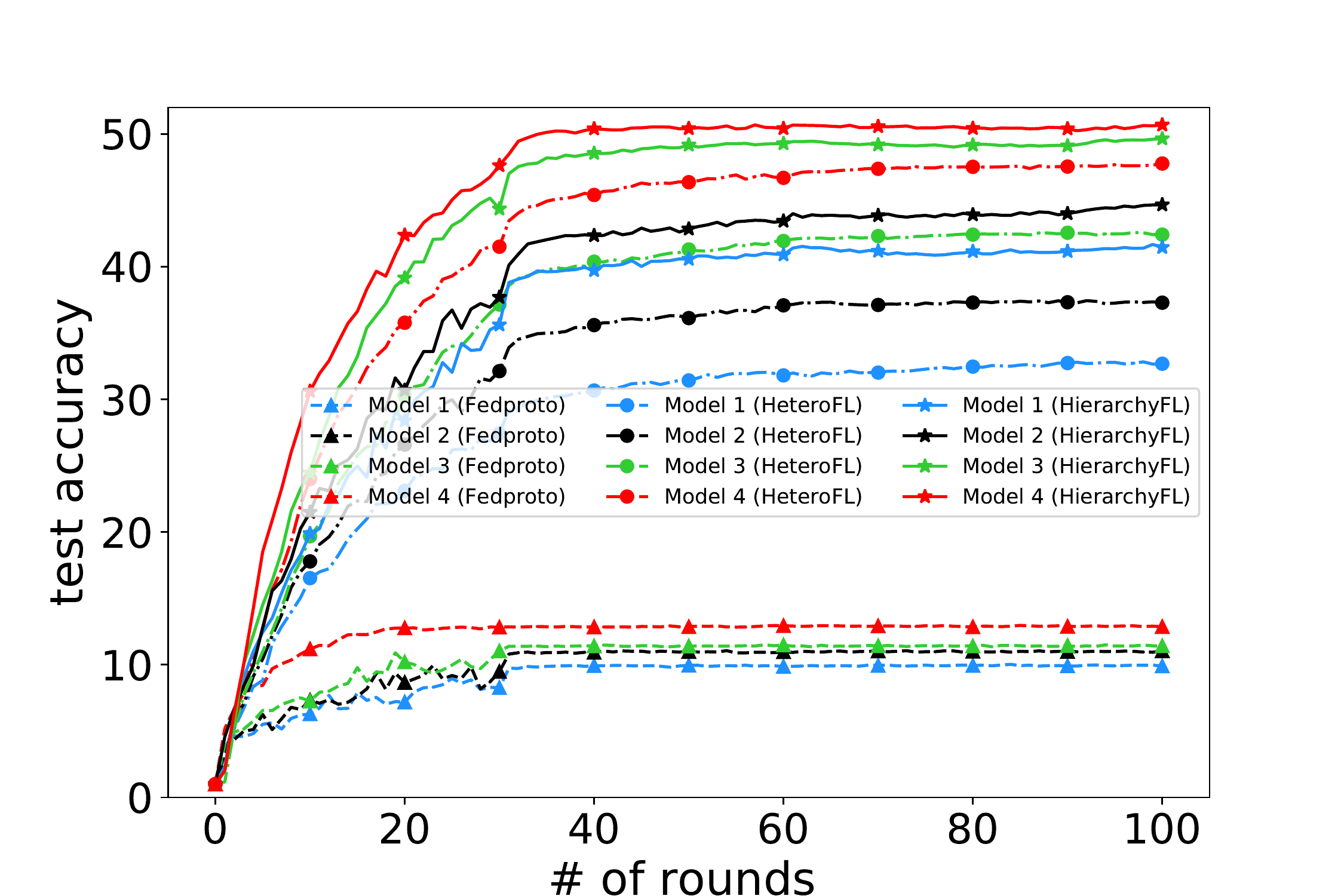}}
\subfigure[\scriptsize{ CIFAR100 ($\alpha=0.5$) w/ 40 devices}]{\includegraphics[width=0.24\linewidth]{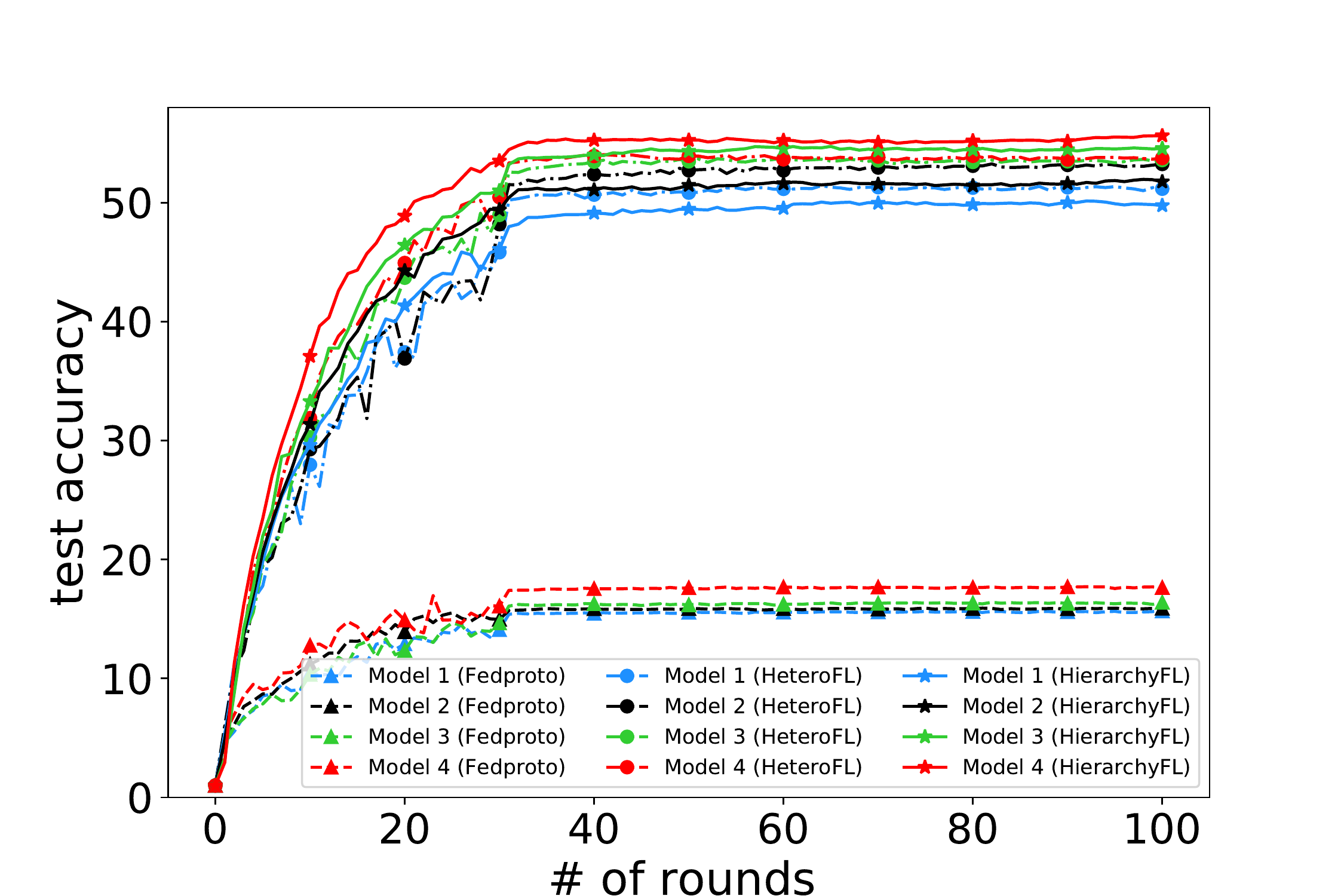}}
    \subfigure[\scriptsize{ CIFAR100 ($\alpha=1$) w/ 40 devices}]{\includegraphics[width=0.24\linewidth]{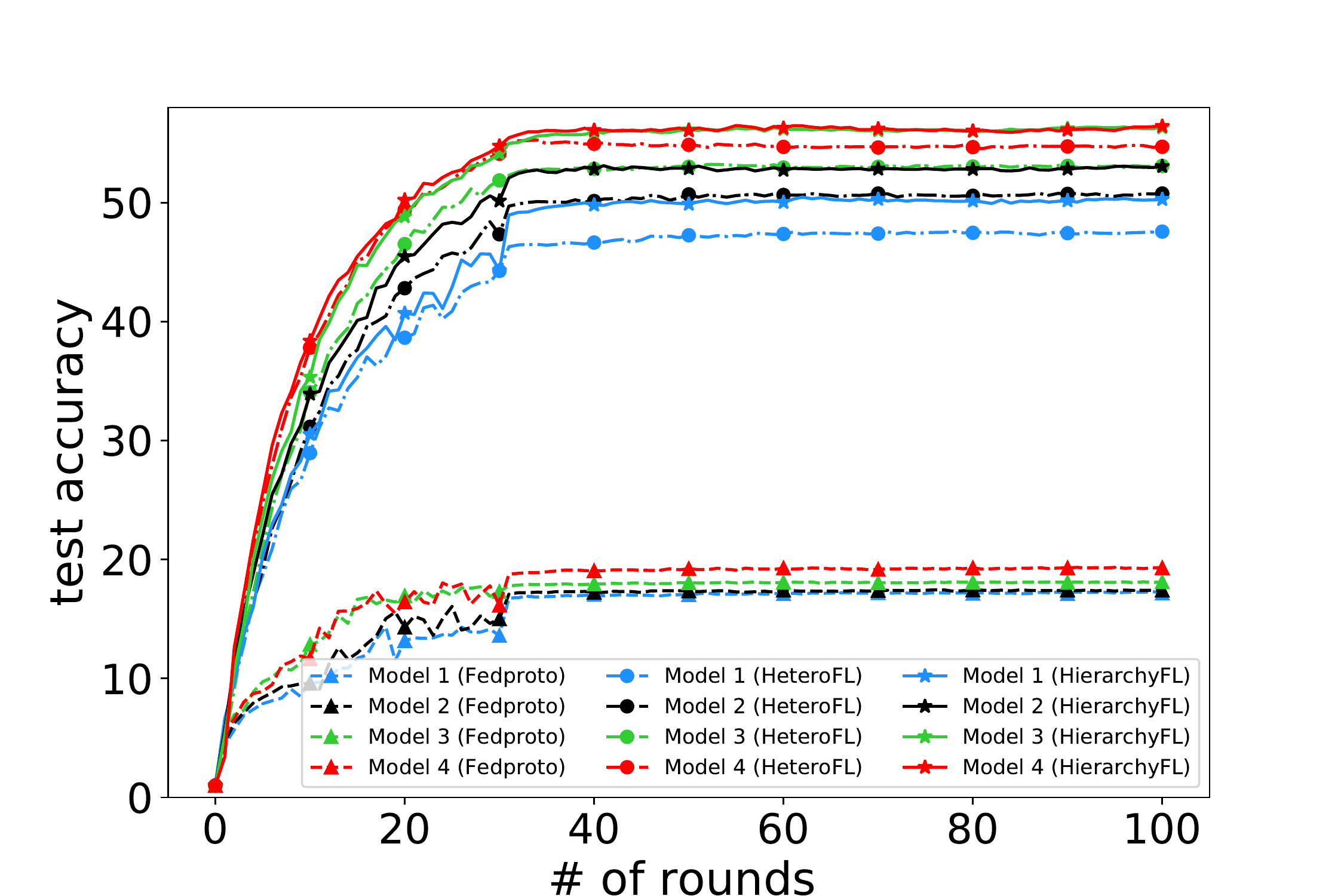}}
\subfigure[\scriptsize{  CIFAR100 (IID) w/ 40 devices}]{\includegraphics[width=0.24\linewidth]{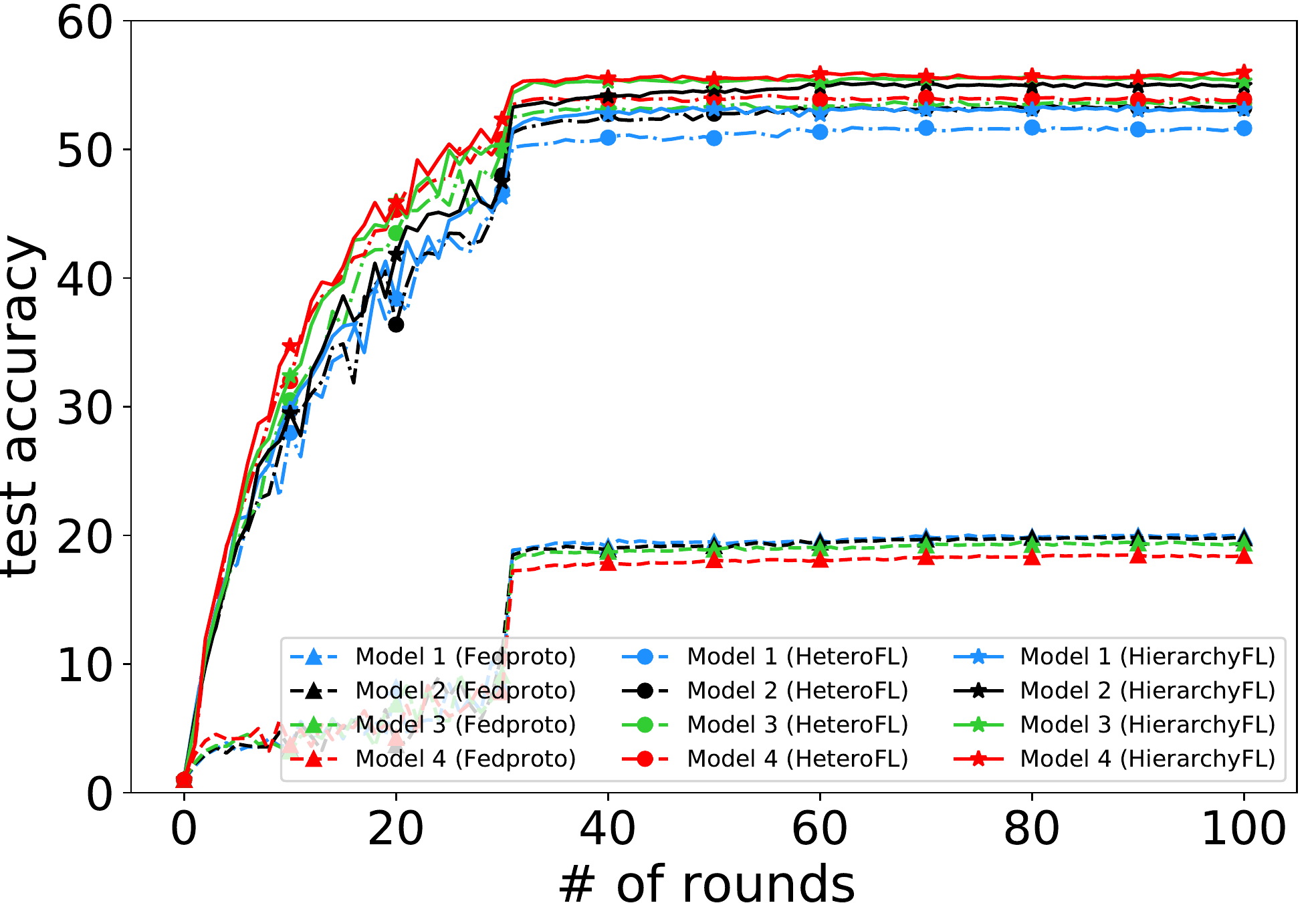}}

    \subfigure[\scriptsize{  CIFAR10 ($\alpha=0.1$) w/ 60 devices}]{\includegraphics[width=0.24\linewidth]{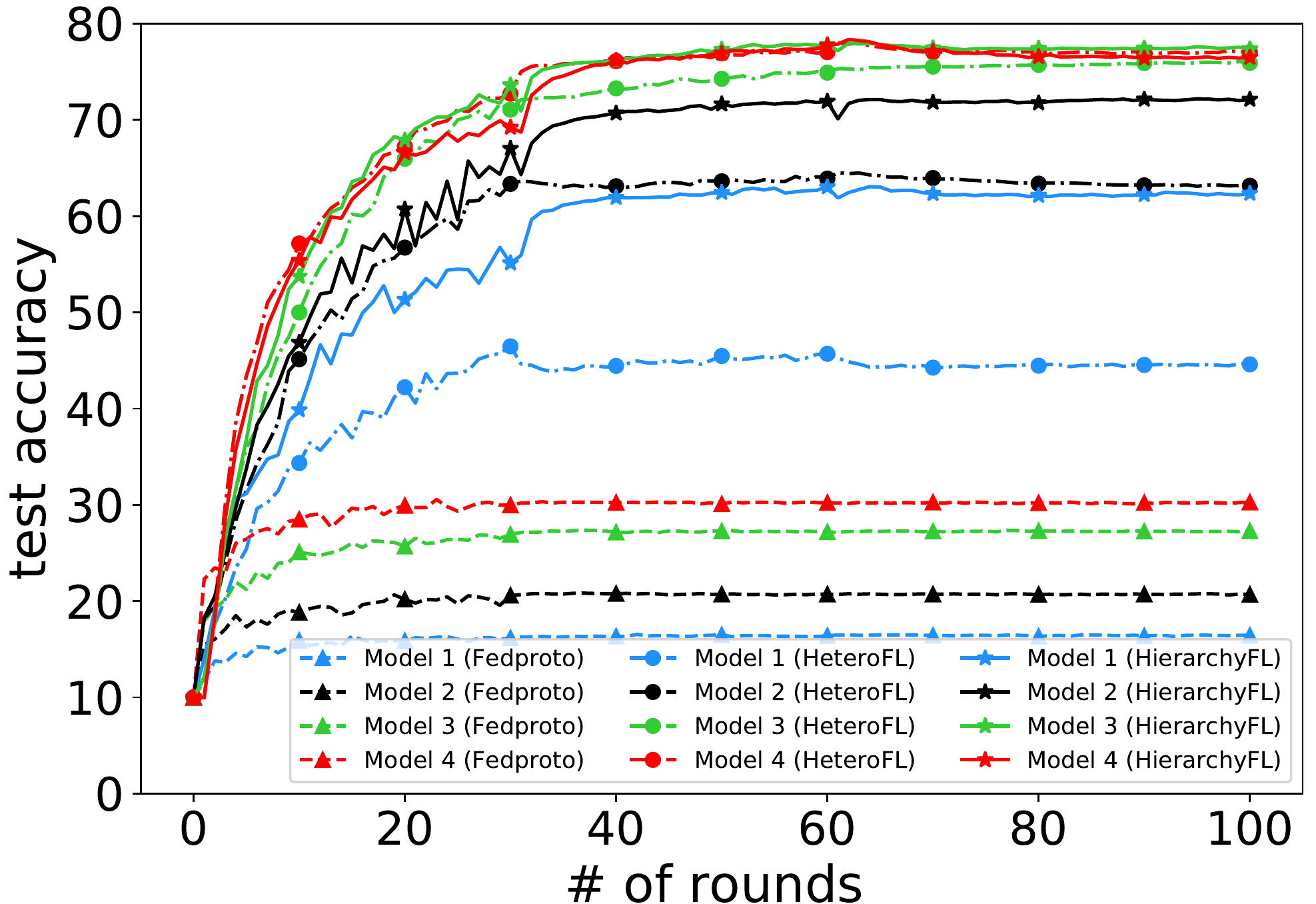}}
\subfigure[\scriptsize{ CIFAR10 ($\alpha=0.5$) w/ 60 devices}]{\includegraphics[width=0.24\linewidth]{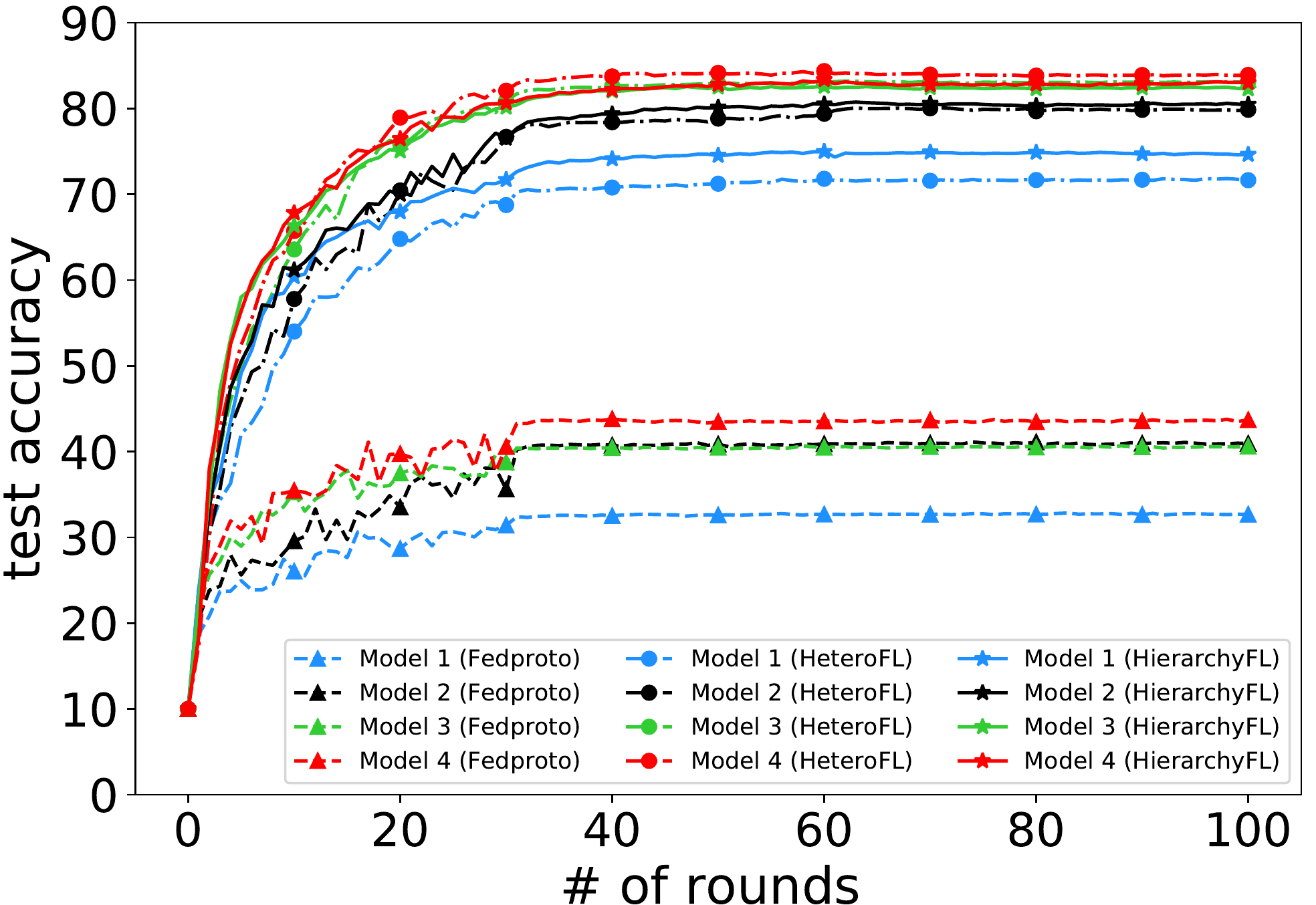}}
    \subfigure[\scriptsize{ CIFAR10 ($\alpha=1$) w/ 60 devices}]{\includegraphics[width=0.24\linewidth]{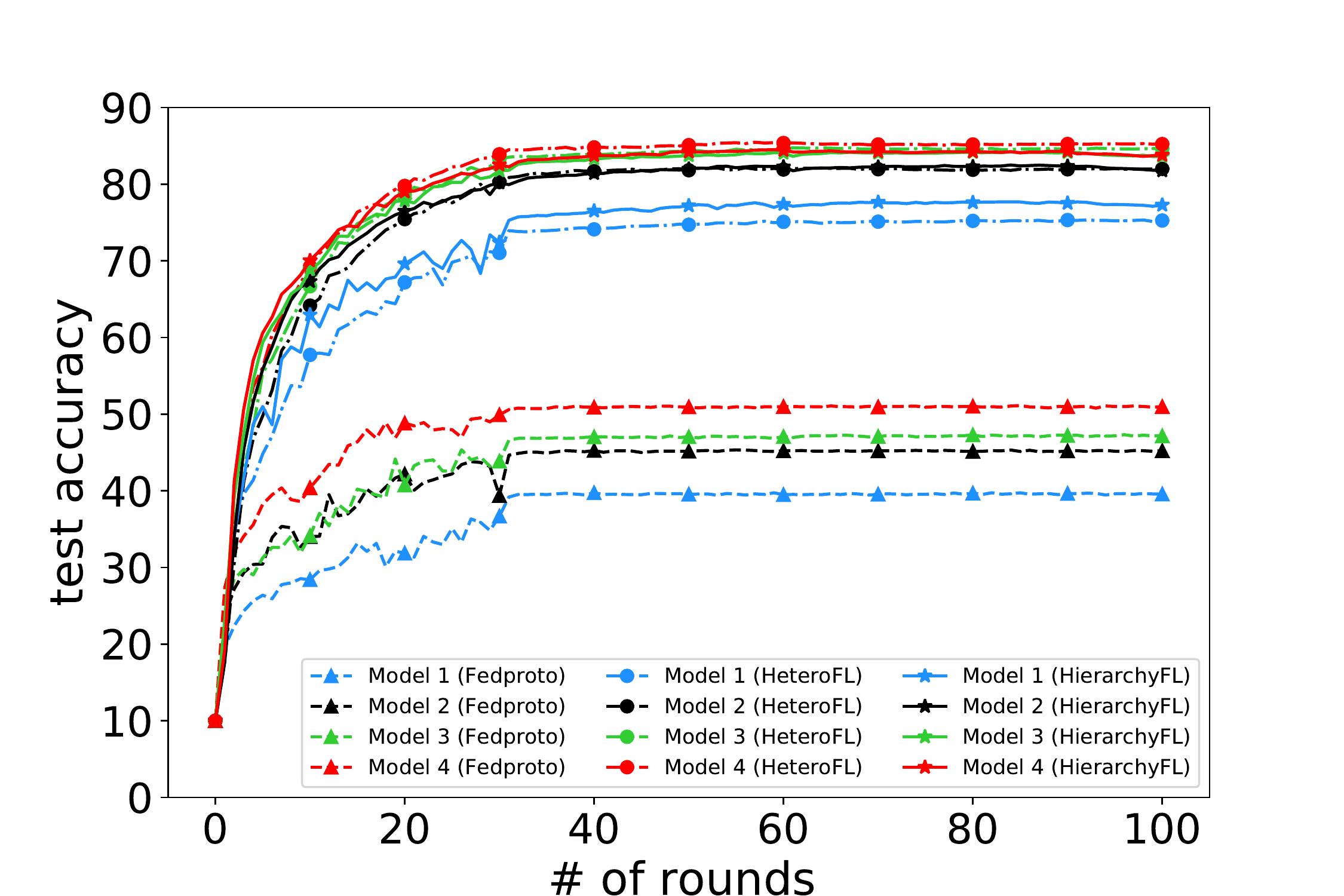}}
\subfigure[\scriptsize{ CIFAR10 (IID) w/ 60 devices}]{\includegraphics[width=0.24\linewidth]{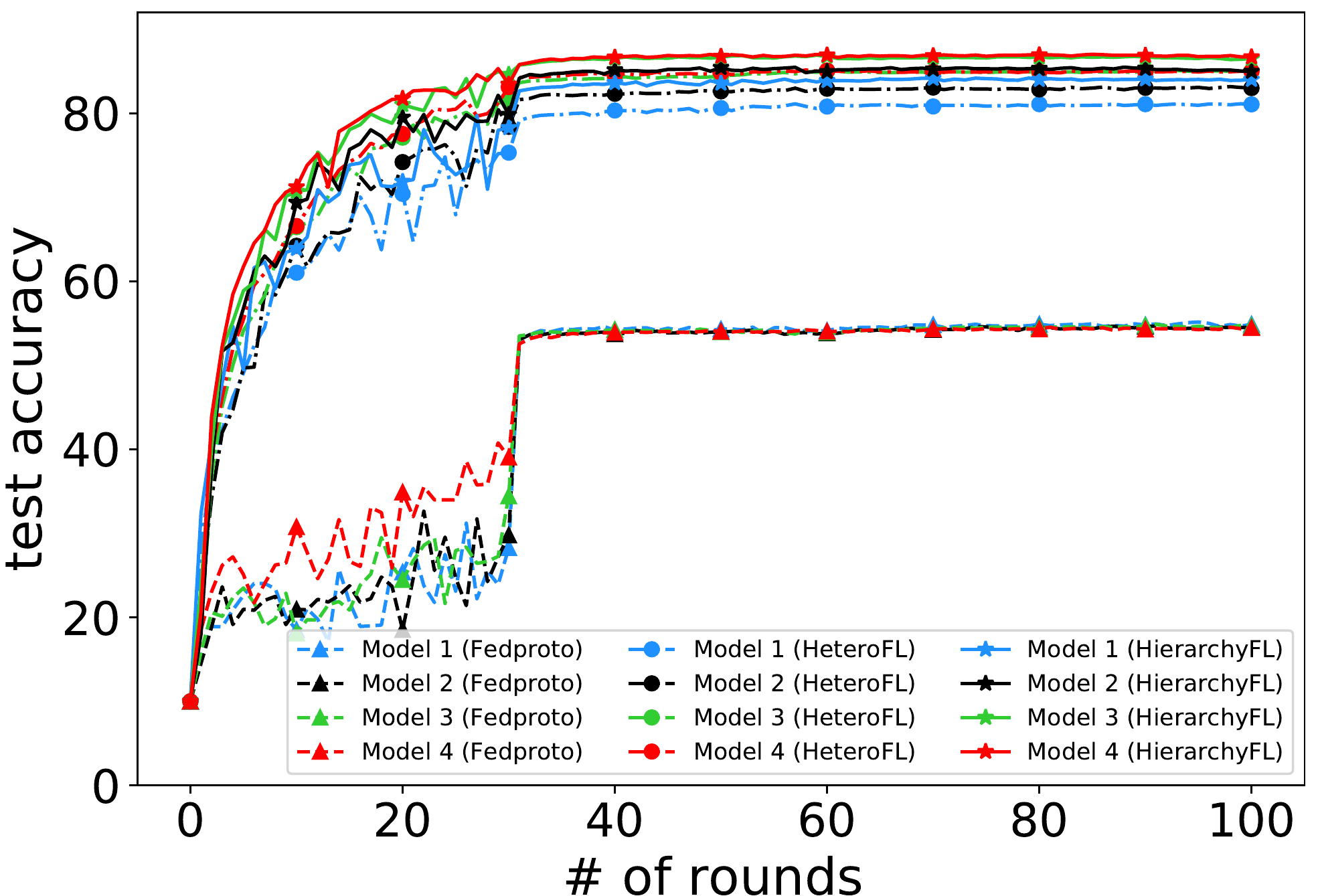}}
    \subfigure[\scriptsize{ CIFAR100 ($\alpha=0.1$) w/ 60 devices}]{\includegraphics[width=0.24\linewidth]{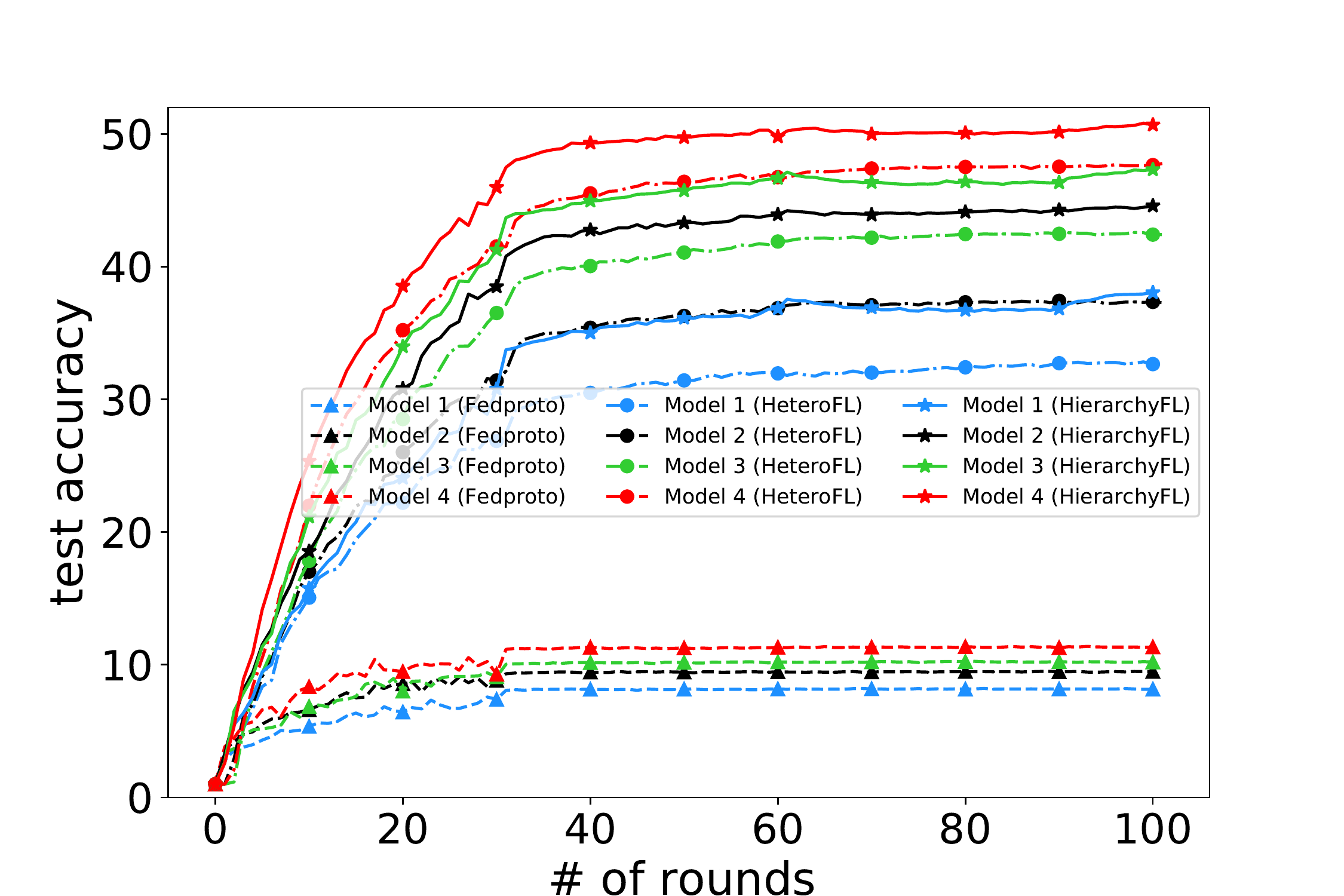}}
\subfigure[\scriptsize{ CIFAR100 ($\alpha=0.5$) w/ 60 devices}]{\includegraphics[width=0.24\linewidth]{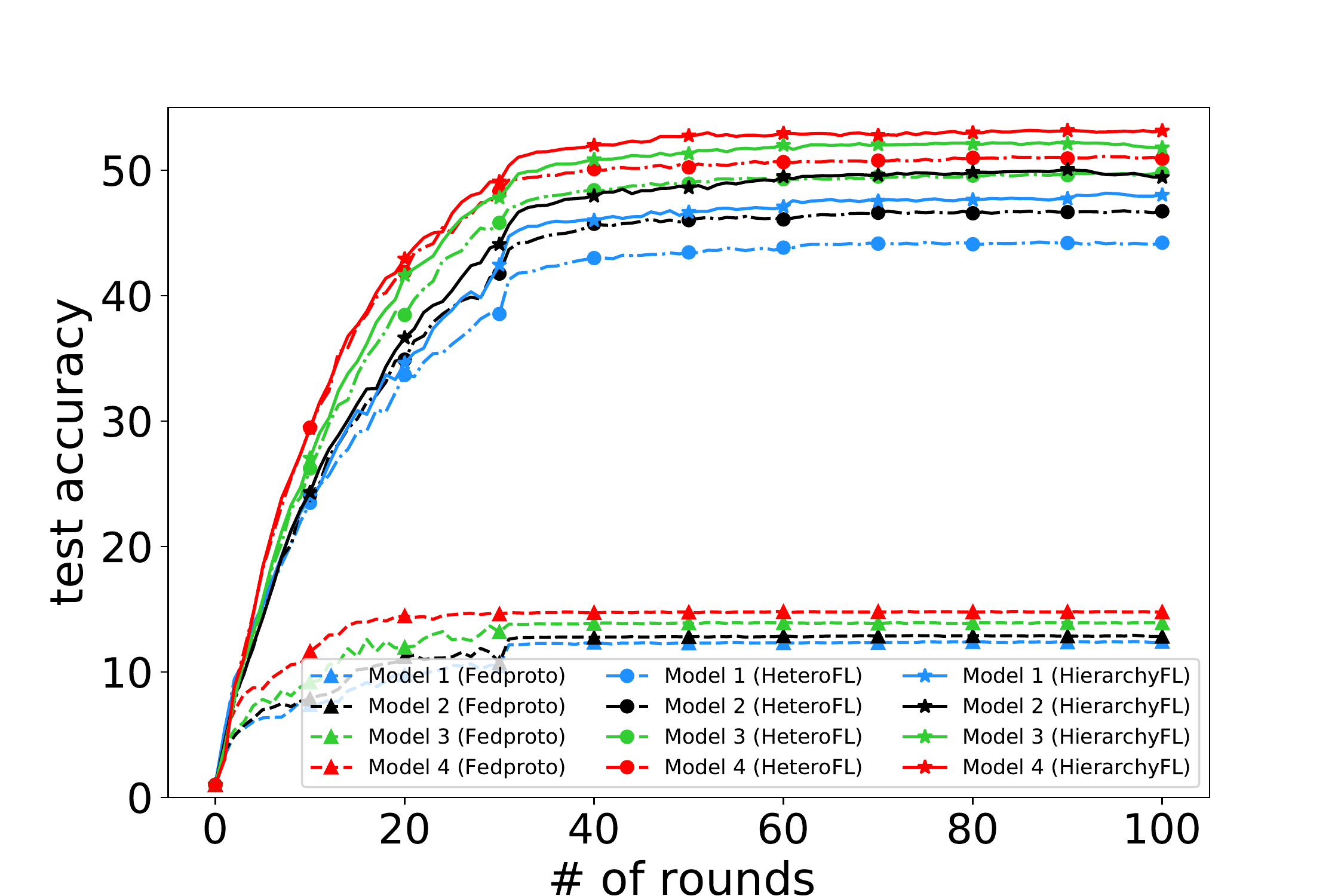}}
    \subfigure[\scriptsize{ CIFAR100 ($\alpha=1$) w/ 60 devices}]{\includegraphics[width=0.24\linewidth]{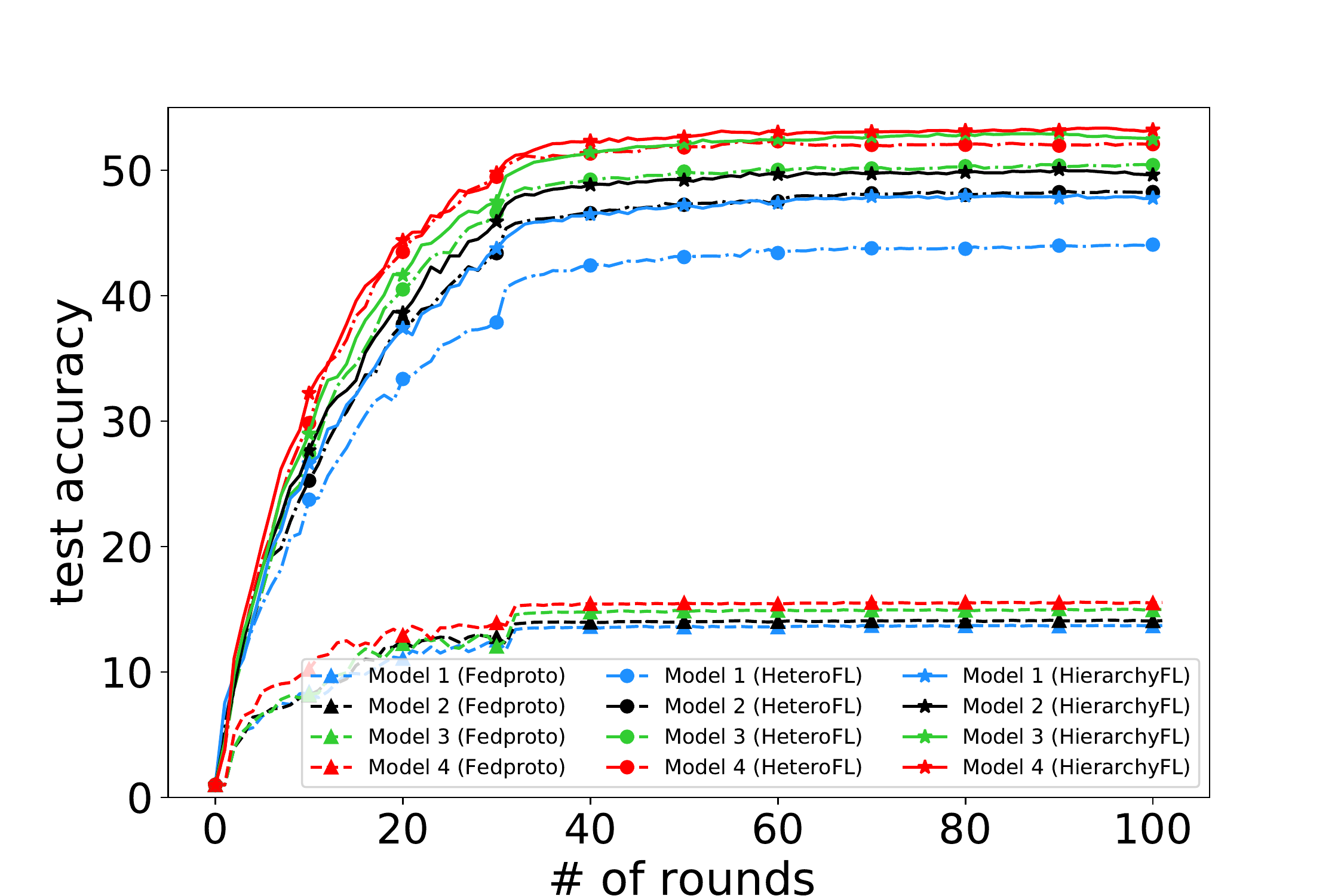}}
\subfigure[\scriptsize{ CIFAR100 (IID) w/ 60 devices}]{\includegraphics[width=0.24\linewidth]{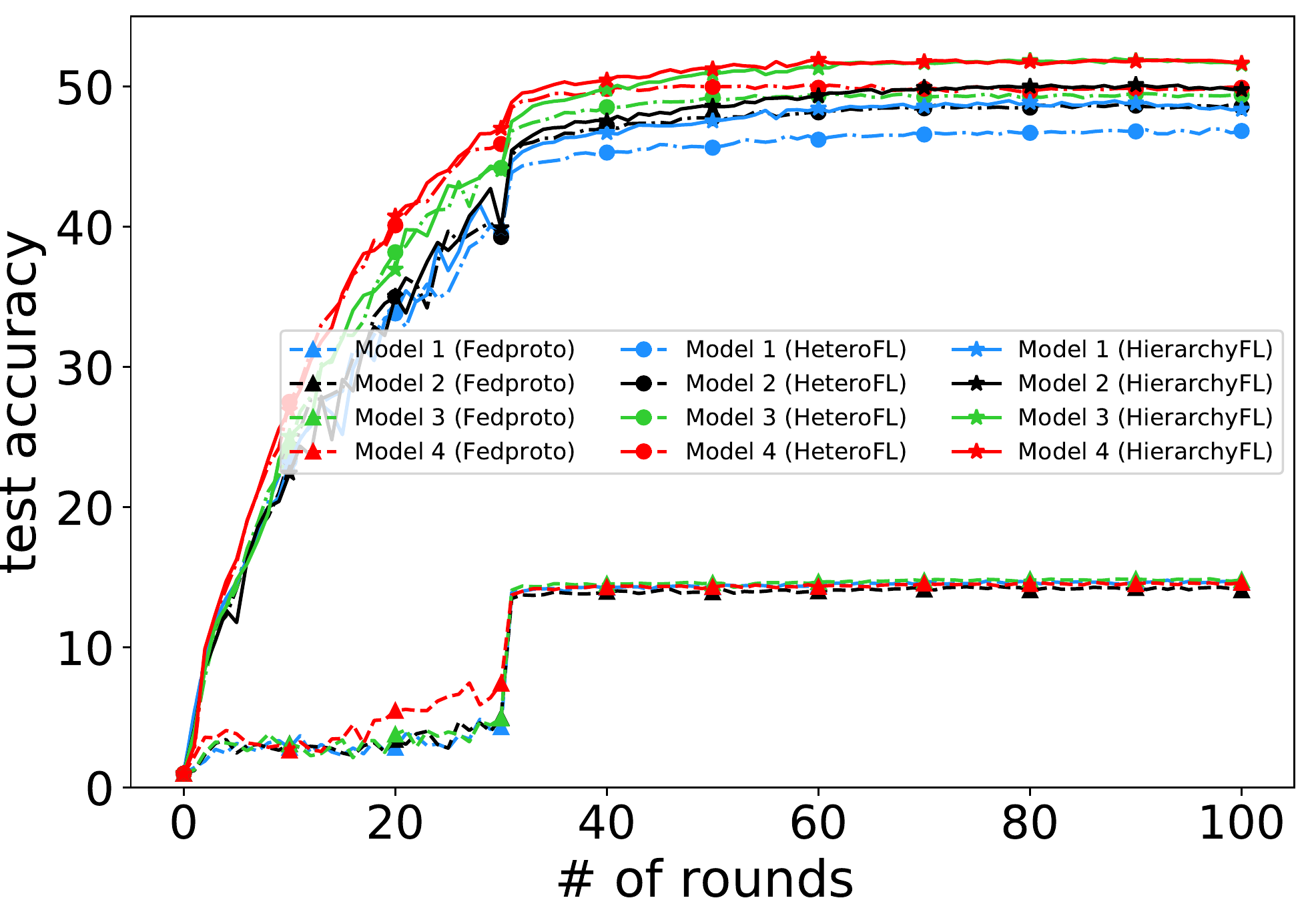}}
\caption{Test accuracy comparison of AIoT systems for both 
 IID and non-IID scenarios with different number of  devices}
\label{sca_result} 
\vspace{-0.2in}
\end{center}
\end{figure*}

To evaluate 
the effectiveness of our proposed  method,  
we developed the  HierarchyFL  framework 
using  PyTorch (version 1.4.0). We 
 set the random seed to  1234, which can 
 mitigate the influence of randomness within the experimental results. Note that in the training procedure, we assumed that all the AIoT devices are involved  in each FL communcation
 round.
We set the mini-batch size of both hierarchical self-distillation and other heterogeneous FL methods to 32,  where the number of local training epochs and the initial learning rate are set to 5 and 0.05, respectively. To ensure the stability of the training process, the learning rate adaptively dropped 0.1 in every 30 epochs. 
The initial learning rate  of HierarchyFL is set to 0.05. Based on the adaptive drop of the learning rate,  it  drops by 0.1 in every 30  epochs. 
Our experimental environment is based on a   Windows personal computer, which is deployed with an Intel i9 CPU, 32GB memory, and  GTX3090. 
The subsequent experiments are conducted to answer these three research questions.

\textbf{RQ1: (Superiority of HierarchyFL)}: 
What are the advantages of HierarchyFL compared with 
state-of-the-art heterogeneous federated learning methods?

\textbf{RQ2: (Scalability of HierarchyFL)}:
What is the impact of the number of AIoT devices for knowledge sharing in  HierarchyFL?

\textbf{RQ3: (Benefits  of Ensemble Library)}: How 
much can HierarchyFL  benefit from the ensemble library during hierarchical self-distillation?
\subsection{Experimental Settings}\label{exp_set}
\subsubsection{Dataset Setting}

To show the effectiveness of our HierarchyFL method,  we considered two training datasets (i.e., CIFAR10 and CIFAR100 \cite{CIFAR}), where both of them involve 50000 training images and 10000 testing images, respectively. For each 
dataset, we investigated both Independent and Identically Distributed (IID) and non-IID  distributions for subsequent experiments. 
% Note that each dataset is set as both IID and various Non-independent and identically distributed (non-IID) scenarios for the subsequent experiments. 
For  non-IID scenarios, we constructed local training sets using heterogeneous data splits with a Dirichlet distribution, which is the same as Fedproto \cite{fedproto}. In the Dirichlet distribution, the value of $\alpha$ means the distribution degree of non-IID, where a smaller $\alpha$ represents a higher distribution degree of non-IID.
% To show  the effectiveness of our approach, we consider two kinds of data settings with IID and various Non-independent and identically distributed (non-IID). For the non-IID setting,  we construct local training sets using heterogeneous data splits with a Dirichlet distribution as the Fedproto \cite{fedproto}. 
 % In the Dirichlet distribution, the value of $\alpha$ means the distribution degree of non-IID. A smaller $\alpha$  represents higher distribution degree of non-IID. 
% For the  image classification task, we use the CIFAR10 and CIFAR100  as the training datasets \cite{CIFAR}. Both of them have 
% 50000 training images and 10000 test images, respectively. 
To fully use  the  nature image datasets, we use the data augmentation technologies  in our experiment, which are similar to the ones used by HeteroFL \cite{iclr_diao2021}. To simulate the real scenario in  the cloud server,  in our experiment, the  public dataset only accounts for only 1\% of the overall training data in HierarchyFL.

\subsubsection{Model Setting}
We compared
our HierarchyFL with two state-of-the-art model heterogeneous methods, i.e., HeteroFL \cite{iclr_diao2021}, and Fedproto \cite{fedproto}. 
Following the HeteroFL, we set the ResNet-18 \cite{ResNet} as the backbone, and four bottlenecks and  classifiers are followed by each block of ResNet18 as the four newly hierarchy models (i.e.,  Models 1-4 shown in \tabref{test_acc}) to simulate four types of  heterogeneous models in our experiments. Note that under these settings, each hierarchy model can be reused with the same backbone to conduct the model inference.

\subsection{Performance comparison}
To show the superiority of HierarchyFL, we compared our 
approach with the two heterogeneous FL methods with four
data settings. \tabref{test_acc} presents the comparison results for two image datasets.  
In this table, columns 2-4 present the name of three heterogeneous FL methods, where each column has four sub-columns to represent the various 
training  data distribution of all local AIoT devices.
Note that  the non-IID setting we used in this table is the Dirichlet distribution with various $\alpha$ settings (i.e., 0.1, 0.5, 1.0).
% Columns 4-7 denote the experimental results for various hierarchy models, respectively. 

From this table,  we can realize that  HierarchyFL obtains the best inference performance in 30 out of 32 scenarios. 
For example, compared with
Fedproto, HierarchyFL achieves more than 25\%
improvement with  $\alpha = 0.1$ for Model 1.  Furthermore, we can find that our HierarchyFL method always outperforms its counterparts in non-IID settings. For instance,
HierarchyFL outperforms HeteroFL by  13\%  in  $\alpha = 0.1$ training data set on average.
% In other words, the hierarchical self-distillation by the ensemble library can enhance the inference capabilities of all kinds of hierarchy models, especially in  non-IID scenarios. 
In other words,  the hierarchical self-distillation can effectively improve the inference capabilities of all the hierarchy models with the assistance of the ensemble library. 
This is mainly because HierarchyFL fully considers the diversity of each hierarchy model knowledge, which can be used in   self-distillation by learning a  wise ensemble library to effectively enhance the model inference performance of each hierarchy model in an AIoT system. 

\subsection{Discussion}\label{Discussion}
\subsubsection{Scalability Analysis}

Due to the prevalence of large-scale AIoT systems, scalability is essential in the development of HierarchyFL.
% By using the same experimental settings in Section~\ref{exp_set}, 
\figref{sca_result} shows the inference comparison of
AIoT systems  under both IID and non-IID scenarios with both 40 and 60 devices, where the experimental settings are the same as that in Section~\ref{exp_set}. 

As shown in \figref{sca_result},  when large devices participate in the three heterogeneous federated learning methods, the superiority of HierarchyFL  becomes more obvious than the other two methods.
 Our HierarchyFL method always outperforms its counterparts for the two image datasets under all the training data settings with 40 and 60 devices, where  the star (i.e., HierarchyFL) is always above the circle (i.e., HeteroFL)  and triangle (i.e., Fedproto) in the \figref{sca_result}.

\subsubsection{Ablation Study}
To evaluate the contributions of ensemble library parts in HierarchyFL, we conducted an ablation study on CIFAR100 under data setting $\alpha = 0.1$ with 20 devices, whose results are shown in \tabref{Ablation}.  Column 1 denotes the case without adopting knowledge distillation or only using the layer alignment averaging for each hierarchy model.  Based on our 
HierarchyFL method, columns 2 and 3 present the two cases indicating
whether the ensemble logits and ensemble features
are included, respectively. Columns 4
indicates the  inference  performance of  the global model (i.e., Model 4).

\begin{table}[h]
  \caption{Ablation  results considering  impacts of 
  Ensemble Library}
\centering
\begin{tabular}{ccc|cc}
\hline
Averaging & Ensemble  Logits &  Ensemble  Features & Global Model (\%)    \\ \hline
$\checkmark$                 &      &        & 47.09  \\
$\checkmark$                    & $\checkmark$      &          & 49.04     \\
$\checkmark$                    & $\checkmark$      & $\checkmark$              & \textbf{52.56}  \\ \hline
\end{tabular}

  \label{Ablation}
\end{table}

Note that HeteroFL can be considered as HierarchyFL without its hierarchical self-distillation procedure by the ensemble library.
Unlike HeteroFL, HierarchyFL takes the synergy of  hierarchical self-distillation by the ensemble library into account. 
From  \tabref{Ablation}, we can find that compared with HeteroFL,  the inference of the global model  (i.e., Model 4) can  be
improved from 47.09\% to 52.56\%.
In other words, each element in the ensemble library plays a crucial role in the knowledge sharing of HierarchyFL.

\section{Conclusion}
\label{con}
%背景
% 问题
% 挑战
% 本文贡献
% 实验效果
% 未来工作
Although FL is promising for privacy-preserving collaborative learning, it still suffers from the model heterogeneity problem due to the different hardware resources of all involved AIoT devices. To alleviate the negative impact of this problem, we propose a novel HierarchyFL framework, which enables efficient knowledge sharing on large-scale AIoT systems with various heterogeneous models. By combining self-distillation with our ensemble library, each hierarchical model can intelligently learn from each other, improving its model inference performance. Comprehensive experiments on well-known datasets demonstrate the effectiveness of HierarchyFL in terms of inference performance and scalability.

\end{document}